\documentclass{article} 
\usepackage{iclr2024_conference,times}

\usepackage{amsmath,amsfonts,bm}

\def\eqref#1{equation~\ref{#1}}

\def\1{\bm{1}}

\DeclareMathAlphabet{\mathsfit}{\encodingdefault}{\sfdefault}{m}{sl}
\SetMathAlphabet{\mathsfit}{bold}{\encodingdefault}{\sfdefault}{bx}{n}

\usepackage{hyperref}
\usepackage{url}
\usepackage{booktabs}       
\usepackage{amsfonts}       
\usepackage{nicefrac}       
\usepackage{microtype}      
\usepackage{xcolor}         
\usepackage{tcolorbox}
\definecolor{lightcyan}{rgb}{0.88, 1.0, 1.0}
\colorlet{mythmback}{lightcyan!40!white}
\newtcolorbox{boxEnv}{
colback=mythmback,coltitle=blue,colframe=mythmback,
center,
width=\linewidth,
boxrule=0.5pt,
left=5pt,right=0pt,
top=2pt,bottom=2pt,
before skip=10pt, after skip=10pt
}
\usepackage{longtable}
\usepackage{graphicx}
\usepackage{listings}
\usepackage{multirow} %
\usepackage{makecell}
\usepackage{subcaption}
\usepackage{listings}
\usepackage{wrapfig}
\usepackage{framed}
\usepackage{threeparttable}
\usepackage{pythonhighlight}

\newcommand{\wizardemoji}{\includegraphics[height=1.2\fontcharht\font`\B]{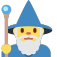}}

\title{\wizardemoji \modelname{}: Empowering Code Large Language Models with Evol-Instruct}

\author{Ziyang Luo$^2$\thanks{\quad Equal contribution. Work done during the internship at Microsoft.}  \quad Can Xu$^1$\footnotemark[1]  \quad Pu Zhao$^1$ \quad Qingfeng Sun$^1$   \quad Xiubo Geng$^1$ \\  {\bf Wenxiang Hu}$^1$ \quad {\bf Chongyang Tao}$^2$ \quad {\bf Jing Ma}$^{2\dagger}$ \quad {\bf Qingwei Lin}$^1$ \quad{\bf Daxin Jiang}$^1$\thanks{\quad  Corresponding author.
 }\\
      $^1$Microsoft\thanks{\quad This paper has been accepted to ICLR 2024. Please cite the ICLR version.} \\
      $^2$Hong Kong Baptist University   \\ 
      \texttt{\{cszyluo, majing\}@comp.hkbu.edu.hk, \{caxu,puzhao\}@microsoft.com}\\
      \texttt{\{qins,xigeng,wenxh,chongyang.tao,qlin,djiang\}@microsoft.com}
      }

\newcommand{\name}{\emph{Evol-Instruct}}
\newcommand{\cname}{\emph{Code Evol-Instruct}}
\newcommand{\modelname}{\emph{WizardCoder}}

\iclrfinalcopy 
\begin{document}

\maketitle

\newcommand{\todo}[1]{\textcolor{brown}{{[#1]}}}

\begin{abstract}

Code Large Language Models (Code LLMs), such as StarCoder, have demonstrated remarkable performance in various code-related tasks. However, different from their counterparts in the general language modeling field, the technique of instruction fine-tuning remains relatively under-researched in this domain. In this paper, we present \cname{}, a novel approach that adapts the \name{} method to the realm of code, enhancing Code LLMs to create novel models \modelname{}.
Through comprehensive experiments on five prominent code generation benchmarks, namely HumanEval, HumanEval+, MBPP, DS-1000, and MultiPL-E, our models showcase outstanding performance. They consistently outperform all other open-source Code LLMs by a significant margin. Remarkably, \modelname{} \textit{15B} even surpasses the well-known closed-source LLMs, including Anthropic's Claude and Google's Bard, on the HumanEval and HumanEval+ benchmarks. Additionally, \modelname{} \textit{34B} not only achieves a HumanEval score comparable to GPT3.5 (ChatGPT) but also surpasses it on the HumanEval+ benchmark. Furthermore, our preliminary exploration highlights the pivotal role of instruction complexity in achieving exceptional coding performance.

\end{abstract}

\begin{figure}[h!]
     \centering
     \begin{subfigure}[b]{\textwidth}
         \centering
         \includegraphics[width=0.95\textwidth]{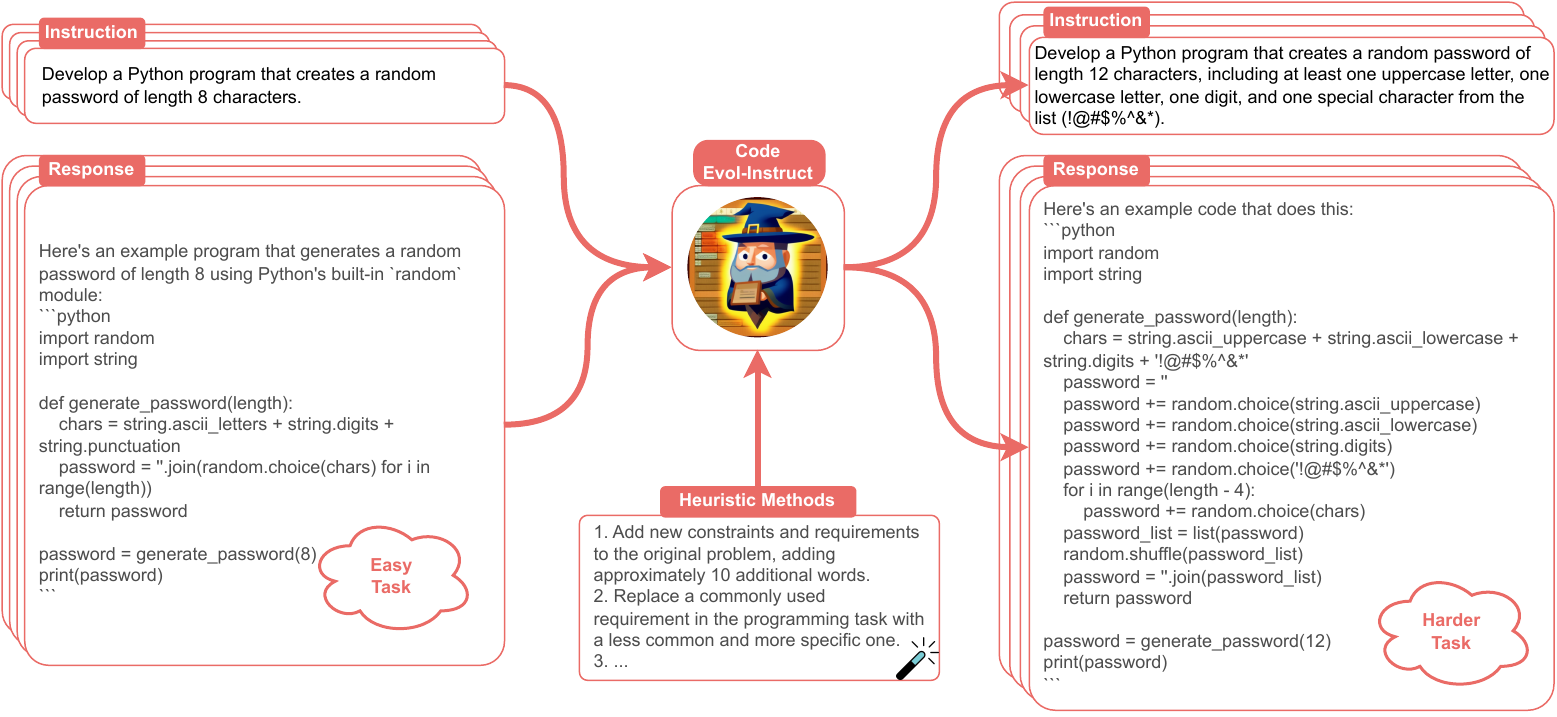}
     \end{subfigure}
     \begin{subfigure}[b]{\textwidth}
         \centering
         \includegraphics[width=0.72\textwidth]{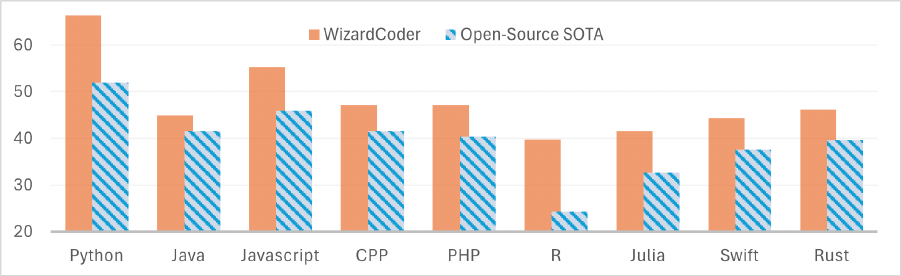}
     \end{subfigure}
        \caption{An illustration of our novel \cname{} and the superior pass@1 performance of our \modelname{} \textit{34B}, outperforming the open-source SOTA (CodeLlama-34B-Series as of the date before August 24, 2023) by a large margin in 9 different programming languages. The Python score is the mean between HumanEval and MBPP.}
        \label{fig:compare_with_sota}
\end{figure}

\section{Introduction}

Recently, Large Language Models (LLMs)~\citep{GPT3,GPT4,PaLM,palm2,Chinchilla,gopher,GLM-130B,opt,llama} have garnered immense attention and demonstrated impressive success. Notably, OpenAI's GPT3.5 (ChatGPT) stands out as a prominent example. These models, through extensive pre-training on vast internet data and fine-tuning with detailed instruction data~\citep{DBLP:conf/nips/Ouyang0JAWMZASR22}, have achieved state-of-the-art (SOTA) zero-shot performance across diverse NLP tasks. This trend also extends to the realm of code understanding and generation, where a multitude of Code LLMs have emerged~\citep{codex,AlphaCode,incoder,codegen,CodeGeeX,codet5,CodeT5+,li2023starcoder,codegen2,codellama}. These models, pre-trained on substantial code data, excel in various code-related tasks, consistently delivering impressive performance.

In contrast to most previous Code LLMs that primarily focus on the pre-training process, there has been limited exploration of fine-grained instruction tuning in the code domain. The introduction of instruction tuning was initially designed to enhance the generalization capabilities of LMs across different tasks via multitask training~\citep{t5,DBLP:conf/iclr/WeiBZGYLDDL22,flan-t5,ExT5,T0,ZeroPrompt,UnifiedQA}. OpenAI's InstructGPT~\citep{DBLP:conf/nips/Ouyang0JAWMZASR22}, for instance, involved soliciting human annotators to provide explicit instructions to ensure alignment with users' intentions. Similarly, recent works such as Alpaca~\citep{alpaca} employed the self-instruct~\citep{wang2022self} method, where GPT3.5 (ChatGPT) generated the instruction data. Vicuna~\citep{vicuna2023} utilized user-shared conversations collected from ShareGPT.com. 
WizardLM~\citep{xu2023wizardlm} introduces the \name{} method, which involves evolving existing general instruction data to generate more complex and diverse datasets. Drawing inspiration from these previous works in the general domain, our work, \cname{}, is specifically tailored to the coding domain's distinctive characteristics.

In this study, we aim to enhance the capabilities of the SOTA open-source Code LLMs (i.e., StarCoder and CodeLlama), by introducing our novel \cname{}. The motivation of this fine-grained instruction-tuning method in the code domain is to automatically increase the complexity of code instruction data, so as to make the best of the internal coding ability of the Code LLMs.
Our \cname{} incorporates several novel methods, including heuristics tailored to coding task features, adversarial sample heuristics, time/space complexity requirements, and evolving stop controls.
The whole process includes two steps: initially, we apply our \cname{} to evolve basic code instruction data, specifically Code Alpaca~\citep{codealpaca}. Subsequently, we fine-tune StarCoder and CodeLlama using our newly generated code instruction-following training set, resulting in our \modelname{} models.

Figure~\ref{fig:compare_with_sota} and the experimental results obtained from five code generation benchmarks, namely HumanEval~\citep{humeval}, HumanEval+~\citep{humanevalp}, MBPP~\citep{MBPP}, DS-100~\citep{DS1000}, and MultiPL-E~\citep{multipl_e}, demonstrate that our \modelname{} models outperform all other open-source Code LLMs (before August 24, 2023), achieving state-of-the-art (SOTA) performance. Remarkably, our \modelname{} \textit{15B} even surpasses well-known Anthropic's Claude and Google's Bard in terms of pass rates on HumanEval and HumanEval+. Furthermore, \modelname{} \textit{34B} not only achieves a HumanEval score comparable to GPT3.5 (ChatGPT) but also surpasses it on the HumanEval+ benchmark. Beyond this, our preliminary studies indicate that the complexity of instructions is the key to achieving exceptional coding performance.

The contributions of this work can be summarized as follows:
\begin{itemize}
    \item We introduce \cname{}, a novel instruction fine-tuning approach for code, which enhances the performance of the open-source Code LLMs by a large margin.
    \item We develop \modelname{} models, which surpass all other open-source Code LLMs by a substantial margin in coding tasks. Notably, the 15B version even outperforms the well-known closed-source LLMs, such as Claude, and Bard. The 34B version achieves a HumanEval score comparable to GPT3.5 (ChatGPT) and surpasses it on the HumanEval+ benchmark.
    \item We conduct a preliminary study highlighting the pivotal role of instruction complexity in achieving exceptional coding performance.
\end{itemize}

\section{Related Work}

\paragraph{Large Language Models.} Recently, LLMs have demonstrated remarkable achievements across a broad spectrum of tasks. Prominent tech companies have made significant strides in developing highly proficient LLMs. These include OpenAI's GPT3\&4~\citep{GPT3,GPT4}, Google's PaLM~\citep{PaLM,palm2}, and Bard\footnote{\url{https://bard.google.com/}}, DeepMind's Chinchilla~\citep{Chinchilla}, and Gopher~\citep{gopher}, as well as Anthropic's Claude\footnote{\url{https://www.anthropic.com/index/introducing-claude}}. However, it is important to note that these models are closed-source and can only be accessed through specific APIs or may not be accessible at all.

The AI community has witnessed the release of several open-source LLMs, where the model weights are made publicly available. EleutherAI has contributed GPT-NeoX-20B~\citep{GPT-NeoX-20B} and GPT-J-6B~\citep{gpt-j}. Google has released UL2-20B~\citep{UL2}. Tsinghua University has introduced GLM-130B~\citep{GLM-130B}. Meta has released OPT~\citep{opt} and LLaMA1\&2~\citep{llama,llama2}. It is worth noting that while these open-source models have made valuable contributions, they generally do not exhibit the same level of performance as their closed-source counterparts.

\paragraph{Large Language Models for Code.} Recent studies have introduced a significant number of LLMs for code-related tasks to address the challenges of code understanding and generation. OpenAI has unveiled Codex~\citep{codex} and Code-Davinci~\citep{Azure}. Google has proposed PaLM-Coder~\citep{PaLM}. They perform outstandingly on the popular code completion benchmarks, like HumanEval~\citep{humeval} and MBPP~\citep{MBPP}. However, these models are closed-source. 

On the other hand, there are several open-source Code LLMs available. Salesforce has introduced CodeGen1\&2~\citep{codegen,codegen2}, CodeT5~\citep{codet5}, and CodeT5+~\citep{CodeT5+}. Tsinghua University has contributed CodeGeeX~\citep{CodeGeeX}, and the BigCode Project has developed StarCoder~\citep{li2023starcoder}. Meta has released the CodeLlama-Series~\citep{codellama}, which achieves open-source SOTA performance on several benchmarks.
The closely related model, CodeLlama-Instruct, refines its performance through the self-instruct method. These models have demonstrated notable advancements in code-related tasks. However, when compared to the SOTA closed-source models, they still lag behind significantly. In contrast to the aforementioned models, our work demonstrates that further training Code LLMs with our \cname{} can substantially enhance performance.

\paragraph{Instruction Fine-Tuning.} 

The primary objective of instruction fine-tuning in its early stages was to enhance the cross-task generalization capabilities of LMs. This was achieved by fine-tuning LMs with a substantial corpus of public NLP tasks. T5~\citep{t5} was among the first models to explore this approach, training on a multitude of supervised text-to-text tasks. Subsequent works such as FLAN~\citep{DBLP:conf/iclr/WeiBZGYLDDL22}, ExT5~\citep{ExT5}, T0~\citep{T0}, and UnifiedQA~\citep{UnifiedQA} further expanded the range of tasks to bolster the overall generalization ability of LMs. Notably, ZeroPrompt~\citep{ZeroPrompt} and FLAN-T5~\citep{flan-t5} pushed the envelope by incorporating thousands of tasks in their training pipelines. Across these studies, a consistent finding emerges: fine-tuning LMs with diverse NLP task instructions yields significant performance improvements when applied to new tasks.

While fine-tuning LMs with diverse NLP tasks has shown promising results, it often falls short in aligning with the intentions of real-world users. OpenAI has pursued a different approach by soliciting human annotators to provide a large corpus of human instructions, encompassing diverse forms and a wide range of task types. Building upon this dataset, OpenAI trained its GPT3~\citep{GPT3} model to create InstructGPT~\citep{DBLP:conf/nips/Ouyang0JAWMZASR22}, which better aligns with users' inputs. This line of development has even led to the impressive work known as GPT3.5 (ChatGPT). However, it is important to note that the dataset and model weights associated with these advancements are not publicly available. Alpaca~\citep{alpaca} takes a different route by adopting the self-instruct method~\citep{wang2022self}, leveraging GPT3.5 (ChatGPT) to generate data for training. Vicuna~\citep{vicuna2023} utilizes user-shared conversations collected from ShareGPT.com to train its models.
WizardLM~\citep{xu2023wizardlm} introduces the \name{} method, which involves evolving existing general instruction data to generate more complex and diverse datasets.  Drawing inspiration from this idea, our work, \cname{}, aligning with the distinctive characteristics of coding domains, is the first instruction fine-tuning method explicitly designed to enhance Code LLMs.
\section{WizardCoder: SOTA Open-Source Code LLM}
In this section, we elaborate on the methodological details of \modelname{}. As illustrated in Figure~\ref{fig:compare_with_sota}, we first adopt our \cname{} to iteratively evolve the Code Alpaca dataset. Subsequently, we fine-tune the pre-trained Code LLMs with the evolved data.

\subsection{Code Evol-Instruct}
Inspired by the \name{} method proposed by WizardLM~\cite{xu2023wizardlm}, this work attempts to automatically enhance the complexity of code instructions, thereby improving the fine-tuning effectiveness of Code LLMs. Diverging from the general domain, our methods are meticulously designed to align with the specific characteristics of coding domains. The evolutionary process introduces the following features:
\begin{enumerate}
    \item Heuristics aligned with coding task features on platforms like LeetCode, strategically increasing the complexity of coding tasks to enhance the model's capabilities.
    \item Introduction of erroneous code as an adversarial sample, inspired by prior research on attacking pre-trained code models~\cite{attackcode,attackcode2}, adds a novel and effective method to escalate task complexity.
    \item Introduction of a heuristic emphasizing time and space complexity leverages insights from previous studies~\cite{constraints}, providing a valuable avenue for improving task complexity.
\end{enumerate}

So, the code evolutionary prompt template is as follows:

\definecolor{beaublue}{rgb}{1.0, 0.46, 0.44}
\newenvironment{myblock}{%
  \begin{tcolorbox}[colback=beaublue!8!white,colframe=beaublue!10!black,title=Prompt for Code Evol-Instruct]
}{%
  \end{tcolorbox}
}

\begin{myblock}
Please increase the difficulty of the given programming test question a bit. \\\\You can increase the difficulty using, but not limited to, the following methods:\\ \{method\}\\\\ \{question\}
\end{myblock}
Here, $\{$question$\}$ represents the current code instruction awaiting evolution, and $\{$method$\}$ is the type of evolution. The five types we used are listed as follows:

\newenvironment{myblock2}{%
  \begin{tcolorbox}[colback=beaublue!8!white,colframe=beaublue!10!black,title=Code Evolution Heuristic Methods]
}{%
  \end{tcolorbox}
}
\begin{myblock2}
Add new constraints and requirements to the original problem, adding approximately 10 additional words.\\\\Replace a commonly used requirement in the programming task with a less common and more specific one.\\\\If the original problem can be solved with only a few logical steps, please add more reasoning steps.\\\\Provide a piece of erroneous code as a reference to increase misdirection.\\\\Propose higher time or space complexity requirements, but please refrain from doing so frequently.
\end{myblock2}

\subsection{Training \modelname{}}
We employ the following procedure to train \modelname{}. Initially, we utilize StarCoder 15B~\citep{li2023starcoder} and CodeLlama-34B-Python~\citep{codellama} as the foundations and proceed to fine-tune them using the code instruction-following training set, which was evolved through \cname{}. The prompt format for fine-tuning is outlined as follows:

\newenvironment{myblock3}{%
  \begin{tcolorbox}[colback=beaublue!8!white,colframe=beaublue!10!black,title=Prompt for Fine-Tuning Format]
}{%
  \end{tcolorbox}
}

\begin{myblock3}
Below is an instruction that describes a task, paired with an input that provides further context. Write a response that appropriately completes the request. \\\\\#\#\# Instruction:\\ \{instruction\}\\ \\\#\#\# Response:
\end{myblock3}
To construct the training dataset, we initialized it with the instruction-following dataset called Code Alpaca\footnote{\url{https://github.com/sahil280114/codealpaca}}. We iteratively employ the \cname{} technique on this dataset consisting of around 20k samples to produce evolved data. After each round of data evolution, we merge the evolved data from all previous rounds with the original dataset to finetune Code LLMs. An external dev set serves as the controlled Evol Stop. If the performance drops, we halt the evolution. In Appendix~\ref{app:sim}, we outline the approach employed to prevent data leakage. Additionally, Appendix~\ref{app:evol_example} showcases some evolved examples for reference.
\section{Experiment}

This section begins by providing a comprehensive overview of the baseline models in our experiments. Subsequently, we present the performance of our models on five code generation benchmarks: HumanEval~\citep{humeval}, HumanEval+~\citep{humanevalp}, MBPP~\citep{MBPP}, DS-1000~\citep{DS1000} and MultiPL-E~\citep{multipl_e}.

\subsection{Baselines}

\paragraph{Closed-Source Models.} Multiple technology companies have successfully developed highly proficient LLMs while choosing not to publicly release them. These models are referred to as closed-source models. For our research, we incorporate a substantial number of these models as our baselines. Specifically, our baselines encompass the following: (i) OpenAI's GPT3.5(ChatGPT)\&GPT4~\citep{GPT4}, Code-Davinci-002~\citep{Azure}, Code-Cushman-001~\citep{Azure}, and Codex~\citep{codex}; (ii) Google's Bard, PaLM 2~\citep{palm2}, PaLM~\citep{PaLM}, and LaMDA~\citep{LaMDA}; (iii) Google DeepMind's AlphaCode~\citep{AlphaCode};(iv) Anthropic's Claude; (v) Huawei's PanguCoder2~\citep{pangucoder2}; and (vi) Meta's Unnatural-CodeLlama-34B~\citep{codellama}.

\paragraph{Open-Source Models.} Several open-source LLMs (OSS) have been made available to the AI community, although their performance generally lags behind the closed-source models a lot. As part of our research, we incorporate a significant number of these open-source models as our baselines. Our baselines encompass the following models: InCoder\cite{incoder}, StarCoder and StarCoder-Plus~\citep{li2023starcoder}, LLaMa1\&2~\citep{llama,llama2}, CodeGen~\citep{codegen}, CodeGeeX~\citep{CodeGeeX}, CodeT5+\citep{CodeT5+}, and CodeLlama~\citep{codellama}. In addition, we also include several models with instructions fine-tuning, including CodeLlama-Instruct~\citep{codellama}, OctoCoder~\citep{octocoder}, InstructCodeT5+~\citep{CodeT5+}, Instruct-Codegen-16B,\footnote{\url{https://huggingface.co/sahil2801/instruct-codegen-16B}} Guanaco-65B~\citep{guanaco}, Falcon-40B-Instruct~\citep{falcon} and Vicuna-13B~\citep{vicuna2023}. More details can be found in the Appendix~\ref{app:baselines}.

\subsection{Implementation Details}

The StarCoder and CodeLlama-34B-Python serve as our basic foundation models. OpenAI's gpt3.5-turbo is used to evolve the dataset and generate responses. The evolved dataset consists of approximately 78k samples. To fine-tune the basic models, we employ specific configurations, including a batch size of 512, a sequence length of 2048, 200 fine-tuning steps, 30 warmup steps, a learning rate of 2e-5, a Cosine learning rate scheduler, and fp16 mixed precision.

\begin{figure}
\centering
     \includegraphics[width=0.78\textwidth]{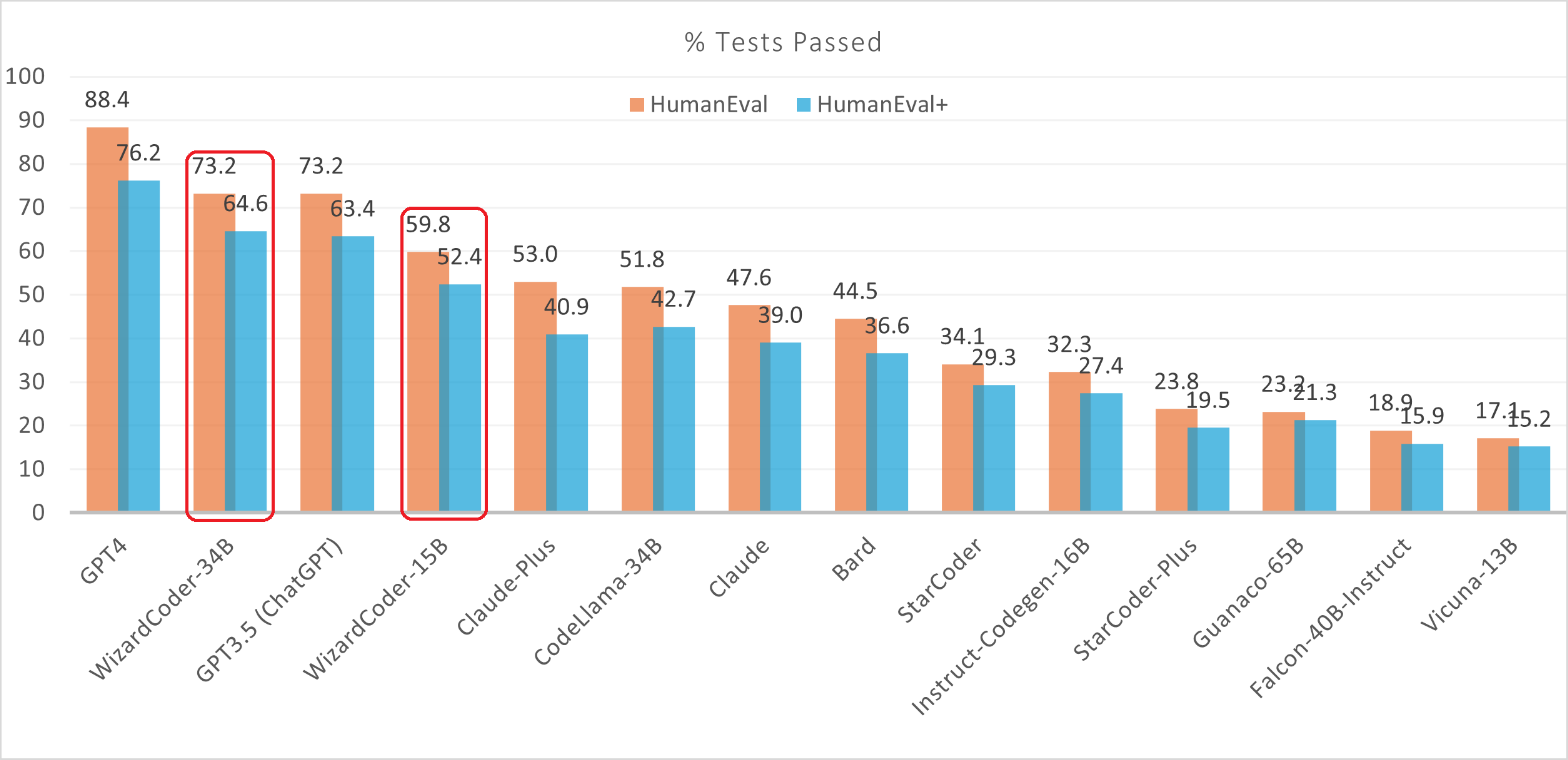}
     \caption{The percentage of pass rates on the HumanEval and HumanEval+ with a single attempt (greedy decoding), following the EvalPlus leaderboard~\citep{humanevalp}.}
     \label{fig:pass1}
\end{figure}

\begin{table}[h!]
    \scriptsize
    \centering
    \caption{Results of pass@1(\%) on HumanEval and MBPP. We follow the previous works~\citep{humeval} to generate n=200 samples to estimate the pass@1 score of our \modelname{} models with the same set of hyper-parameters: temperate=0.2, and top\_p=0.95. *: our reproduced results.}
    \begin{tabular}{lccc}
        \toprule
        \textbf{Model} & \textbf{Params} & \textbf{HumanEval} & \textbf{MBPP}\\
        \midrule
        \multicolumn{4}{c}{Closed-source models}\\
        \midrule
        LaMDA~\citep{LaMDA} & 137B & 14.0 & -\\
        AlphaCode~\citep{AlphaCode} & 1.1B & 17.1 & -\\
        PaLM~\citep{PaLM} & 540B & 26.2 & 36.8\\
        PaLM-Coder~\citep{PaLM} & 540B & 36.0 & 47.0\\
        PaLM 2-S~\citep{palm2} & Unknown & 37.6 & 50.0\\
        Codex~\citep{codex} & 2.5B & 21.4 & -\\
        Codex~\citep{codex} & 12B & 28.8 & -\\
        Code-Cushman-001~\citep{Azure} & Unknown & 33.5 & 45.9\\
        Code-Davinci-002~\citep{Azure} & Unknown & 47.0 & 58.1\\
        GPT-3.5 (ChatGPT)~\citep{GPT4} & Unknown & 48.1 & 52.2\\
        PanguCoder2~\citep{pangucoder2} & 15B & 61.6 & -\\
        Unnatural-CodeLlama~\citep{codellama} & 34B & 62.2 & 61.2\\
        GPT-4~\citep{GPT4} & Unknown & 67.0 & -\\
        \midrule
        \multicolumn{4}{c}{Open-source models}\\
        \midrule
        Llama~\citep{llama} & 65B & 23.7 & 37.7\\
        Llama2~\citep{llama2} & 70B & 29.9 & 45.0\\
        CodeGen-Mono~\citep{codegen} & 16B & 29.3 & 35.3\\
        CodeGeeX~\citep{CodeGeeX} & 13B & 22.9 & 24.4\\
        StarCoder~\citep{li2023starcoder} & 15B & 33.6 & 43.6$^*$\\
        CodeT5+~\citep{CodeT5+} & 16B & 30.9 & - \\
        InstructCodeT5+~\citep{CodeT5+} & 16B & 35.0 & -\\
        OctoCoder~\citep{octocoder} & 15B & 46.2 & -\\
        CodeLlama~\citep{codellama} & 34B & 48.8 & 55.0\\
        CodeLlama-Python~\citep{codellama} & 34B & 53.7 & 56.2\\
        CodeLlama-Instruct~\citep{codellama} & 34B & 41.5 & 57.0\\
        \midrule
        \modelname & 15B & \textbf{57.3} & \textbf{51.8}\\
        \modelname & 34B & \textbf{71.5} & \textbf{61.2}\\
        \bottomrule
    \end{tabular}
    \label{tab:humaneval_mbpp}
\end{table}

\subsection{Evaluation on HumanEval, HumanEval+, and MBPP}

HumanEval~\citep{humeval}, HumanEval+~\citep{humanevalp}, and MBPP~\citep{MBPP} are key benchmarks in the Code LLM field, featuring diverse Python programming problems validated using test cases. HumanEval comprises 164 problems with an average of 9.6 test cases per problem. HumanEval+ expands the test cases significantly to an average of 774.8 per problem. In contrast, MBPP provides 500 test programming problems with three automated test cases each.\footnote{For a fair comparison, we present results for GPT3.5(ChatGPT)\&GPT4 using Eval-Plus with the latest OpenAI's APIs~\citep{humanevalp} (Figure~\ref{fig:pass1}) and OpenAI's report~\citep{GPT4} (Table~\ref{tab:humaneval_mbpp}). Prompt format details are in Appendix~\ref{app:prompt}.}

\paragraph{Comparing with the Closed-Source Models.} Following the same setting of the EvalPlus leaderboard~\citep{humanevalp}. In Figure~\ref{fig:pass1}, we compare our \modelname{} models with the closed-source models, such as GPT4, Claude, and Bard on this leaderboard. Notably, all models generate code solutions for each problem utilizing a single attempt, and the resulting pass rate percentage is reported. To maintain consistency, we employ the same experimental setup by generating answers using greedy decoding and evaluate our \modelname{} models using the provided evaluation codes.

As depicted in Figure~\ref{fig:pass1}, our \modelname{} \textit{34B} attains the second position in this benchmark, surpassing GPT3.5 (ChatGPT, 64.6 vs. 63.4) on HumanEval+. Our 15B version outperforms Claude-Plus (59.8 vs. 53.0) and Bard (59.8 vs. 44.5). Furthermore, our \modelname{} models demonstrate a remarkable superiority over other open-source LLMs that undergo instruction fine-tuning.

\paragraph{Comparing with the Open-Source Models.} In Table~\ref{tab:humaneval_mbpp}, we conduct a comprehensive comparison of our \modelname{} with other open-source models on the HumanEval and MBPP benchmarks. In contrast to the results presented in Figure~\ref{fig:pass1}, we adhere to the approach outlined in previous studies~\cite{humeval} by generating n samples for each problem to estimate the pass@1 score. The findings presented in Table~\ref{tab:humaneval_mbpp} clearly demonstrate that our \modelname{} exhibits a substantial performance advantage over all the open-source models.


\subsection{Evaluation on Multi-Language Coding}

We included comprehensive assessment results across 8 distinct programming languages on the MultiPL-E benchmarks. These languages encompass Java, JavaScript, C++, PHP, R, Julia, Swift, and Rust. The empirical results, as presented in Table~\ref{tab:multipl_e}, distinctly demonstrate the superior performance of our \modelname{} models across all evaluated programming languages, surpassing the SOTA open-source Code LLMs. This underscores the efficacy of our \cname{} method.

\subsection{Evaluation on DS-1000}

The DS-1000 benchmark~\cite{DS1000} comprises 1k distinct data science workflows spanning 7 libraries. It assesses the performance of code generations against test cases and supports two evaluation modes: completion and insertion. In our experiments, we only report insertion scores for models that support. In Table~\ref{tab:ds}, we present pass@1 (n=40) results for each library, along with an overall score.\footnote{Given that this benchmark and its evaluation codes are not designed for the instruction fine-tuned models, we encounter significant challenges in aligning our 34B model with this framework. Moreover, the Codellama-34B base model does not support code insertion. Thus, we only include our 15B model results.} Based on these results, our conclusion is that \modelname{} demonstrates a significant superiority over all other models when tackling data science problems on the DS-1000 benchmark.

\begin{table}[h!]
    \scriptsize
    \centering
    \caption{Results of pass@1(\%) on 8 different programming languages on the MultiPL-E~\citep{multipl_e} benchmarks. All models are evaluated with the same set of hyper-parameters: temperature=0.2, top\_p=0.95, max\_length=512, and n=50.}
    \begin{tabular}{lccccccccc}
        \toprule
        \textbf{Model} & \textbf{Params} & \textbf{Java} & \textbf{Js} & \textbf{CPP} & \textbf{PHP} & \textbf{R} & \textbf{Julia} & \textbf{Swift} & \textbf{Rust}\\
        \midrule
        CodeGen-Multi & 16B & 22.2 & 19.2 & 21.0 & 8.4 & 6.5 & 0 & 1.3 & 4.2\\
        CodeGeeX & 13B & 19.1 & 16.9 & 16.9 & 13.5 & 3.9 & 0.3 & 7.3 & 7.9\\
        Code-Cushman-001 & - & 31.9 & 31.3 & 30.6 & 29.0 & 11.0 & 1.5 & 22.1 & 25.2\\
        StarCoderBase & 15B & 28.5 & 31.7 & 30.6 & 26.8 & 10.2 & 21.1 & 16.7 & 24.5\\
        StarCoder & 15B & 30.2 & 30.8 & 31.6 & 26.1 & 15.5 & 23.0 & 22.7 & 21.8\\
        CodeLlama & 34B & 40.2 & 41.7 & 41.4 & 40.4 & 22.7 & 31.4 & 35.3 & 38.7\\
        CodeLlama-Python & 34B & 39.5 & 44.7 & 39.1 & 39.8 & 22.4 & 31.4 & 34.3 & 39.7 \\
        CodeLlama-Instruct & 34B & 41.5 & 45.9 & 41.5 & 37.0 & 24.3 & 32.7 & 37.6 & 39.3\\
        \midrule
        \modelname{} & 15B & \textbf{35.8} & \textbf{41.9} & \textbf{39.0} & \textbf{39.3} & \textbf{33.5} & \textbf{34.0} & \textbf{33.7} & \textbf{27.1}\\
        \modelname{} & 34B & \textbf{44.9} & \textbf{55.3} & \textbf{47.2} & \textbf{47.2} & \textbf{39.8} & \textbf{41.5} & \textbf{44.3} & \textbf{46.2}\\
        \bottomrule
    \end{tabular}
    \label{tab:multipl_e}
\end{table}
\begin{table}[h!]
    \scriptsize
    \centering
    \caption{Performance of \modelname{} \textit{15B} and baseline models on DS-1000. All models are evaluated with the same set of hyper-parameters: temperature=0.2, top\_p=0.5, max\_length=1024. Scores are average pass@1 accuracy over 40 samples. Matplotlib (plt) task does not have the right context, so insertion and completion scores are identical.}
    \begin{tabular}{lccccccccc}
        \toprule
        \textbf{Format} & \textbf{Model} & \textbf{plt} & \textbf{np} & \textbf{pd} & \textbf{py} & \textbf{scp} & \textbf{sk} & \textbf{tf} & \textbf{All}\\
        \midrule
        & \# of problems: & 155 & 220 & 291 & 68 & 106 & 115 & 45 & 1,000\\
        \midrule
        Completion & InCoder-6B & 28.3 & 4.4 & 3.1 & 4.4 & 2.8 & 2.8 & 3.8 & 7.4\\
        Completion & CodeGen-mono & 31.7 & 10.9 & 3.4 & 7.0 & 9.0 & 10.8 & 15.2 & 11.7\\
        Completion & Code-Cushman-001 & 40.7 & 21.8 & 7.9 & 12.4 & 11.3 & 18.0 & 12.2 & 18.1\\
        Completion & StarCoder & 51.7 & 29.7 & 11.4 & 21.4 & 20.2 & \textbf{29.5} & 24.5 & 26.0\\
        Completion & \modelname & \textbf{55.2} & \textbf{33.6} & \textbf{16.7} & \textbf{26.2} & \textbf{24.2} & 24.9 & \textbf{26.7} & \textbf{29.2}\\
        \midrule
        Insertion & InCoder-6B & 28.3 & 4.6 & 2.9 & 4.4 & 2.8 & 3.1 & 7.8 & 7.5\\
        Insertion & StarCoder & 51.7 & 30.8 & 10.3 & 21.0 & 20.2 & 27.4 & 20.0 & 25.4\\
        Insertion & \modelname & \textbf{55.2} & \textbf{35.1} & \textbf{20.4} & \textbf{30.4} & \textbf{28.9} & \textbf{32.3} & \textbf{37.8} & \textbf{32.8}\\
        \bottomrule
    \end{tabular}
    \label{tab:ds}
\end{table}

\section{Analysis}
\begin{wraptable}{r}{6.5cm}
    \scriptsize
    \centering
    \caption{Different evolution execution models.}
    \begin{tabular}{lccc}
        \toprule
        Base Model & Evol Model & Pass@1\\
        \midrule
        StarCoder-15B & GPT-4 & 62.2\\
        StarCoder-15B & GPT-3.5 & 59.8\\
        StarCoder-15B & CodeLlama & 55.5\\
        CodeLlama-34B & GPT-4 & 73.8\\
        CodeLlama-34B & GPT-3.5 & 73.2\\
        CodeLlama-34B & CodeLlama-34B & 70.1\\
        \bottomrule
    \end{tabular}
    \label{tab:execution}
\end{wraptable}

\paragraph{Evolution Models and Rounds.} In Table~\ref{tab:execution}, GPT-4 replaces GPT-3.5 for evolved rounds, boosting HumanEval Pass@1 scores to 73.8 (34B) and 62.2 (15B). Using OSS CodeLlama-Instruct-34B also proves effective, yielding scores of 70.1 (34B) and 55.5 (15B). Despite GPT-4's superior coding performance (88.4 vs. 73.2), the gain in evolved rounds is not proportional (73.8 vs. 73.2). Conversely, CodeLlama's weaker performance narrows when using \cname{} (73.2 vs. 70.1), highlighting its crucial role. More experiments details are listed in Appendix~\ref{app:diff_evol}. Additionally, Figure~\ref{fig:ablation1} presents results for different data evolution rounds. All models are fine-tuned with 200 steps. Due to the limited size of the dev set of MBPP, we merged the training set and dev set, forming the MBPP-400 dev set. The experiments reveal that the highest pass@1 scores on both the MBPP-400 dev set and the HumanEval are achieved subsequent to three rounds of evolution.


\begin{figure}
     \centering
     \begin{subfigure}[b]{0.42\textwidth}
         \centering
         \includegraphics[width=0.9\textwidth]{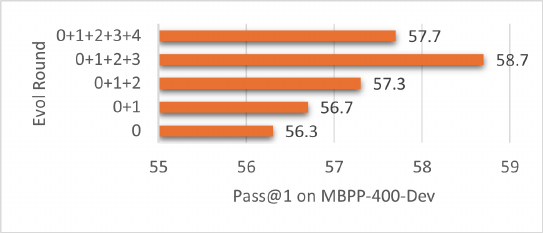}
         \caption{Pass@1 performance on MBPP-400 dev set.}
         \label{fig:mbpp_dev}
     \end{subfigure}
     \hfill
     \begin{subfigure}[b]{0.42\textwidth}
         \centering
         \includegraphics[width=0.9\textwidth]{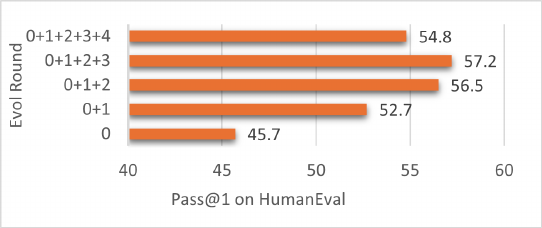}
         \caption{Pass@1 performance on HumanEval.}
         \label{fig:humaneval_dev}
     \end{subfigure}
        \caption{The impact of the number of data evolution rounds.}
        \label{fig:ablation1}
\end{figure}

\paragraph{Complexity and Quantity.}

While the enhanced performance attributed to our \cname{} method has been evident in prior experiments, it remains an open question whether this performance gain is a result of an increase in the number of samples or tokens. During the evolution, each round includes more samples, and the introduction of more complex instructions inevitably leads to an increase in tokens within the training data. To address this question, we fine-tune the models using only the specific round data separately from scratch with a similar number of samples (upper part) or tokens (lower part) in Table~\ref{tab:num}.

\begin{wraptable}{r}{5.5cm}
    \scriptsize
    \centering
    \caption{Analysis of whether the performance gain comes from more tokens.}
    \begin{tabular}{lccc}
        \toprule
        \textbf{Evol} & \textbf{\#Samples} & \textbf{Pass@1}\\
        \midrule
        Round 0 & 20.0k & 45.7\\
        Round 1 & 18.8k & 56.1\\
        Round 2 & 19.7k & 53.0\\
        Round 3 & 19.3k & 54.3\\
        Round 4 & 19.0k & 51.2\\
        \midrule
        \textbf{Evol} & \textbf{\#Tokens} & \textbf{Pass@1}\\
        \midrule
        Round 0 & 2.3M & 44.5\\
        Round 1 & 2.3M & 51.8\\
        Round 2 & 2.3M & 52.4\\
        Round 3 & 2.3M & 50.0\\
        Round 4 & 2.3M & 49.4\\
        \bottomrule
    \end{tabular}
    \label{tab:num}
\end{wraptable}

When each round contains the same number of samples or tokens, the models trained with the seed data still lag behind the evolved rounds. Furthermore, combining data from different rounds leads to the best performance. These results suggest that the primary source of the gain is indeed attributable to our \cname{} method, rather than merely an increase in samples or tokens.

\begin{wrapfigure}{r}{0.4\textwidth}
    \centering
    \vspace{-0.4cm}
    \includegraphics[width=0.25\textwidth]{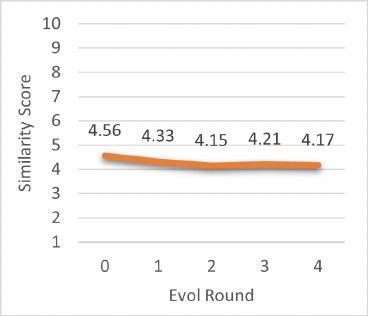}
    \caption{Average similarity scores between HumanEval samples and the top-1 retrieved data, ranging from 1 (completely different) to 10 (identical).}
    \label{fig:sim}
    \vspace{-0.4cm}
\end{wrapfigure}

\paragraph{Complexity and Similarity.} Apart from the quantity analysis, we also investigate whether evolution leads to the inclusion of data more similar to the test set. To address this, we perform an analysis of the HumanEval test set. We employ test samples as queries to retrieve the top-1 sample from each evolved round's training data, utilizing the SOTA embeddings model, gte-large~\citep{gte}. Additionally, we employ GPT4, to provide average similarity scores between the test set and the retrieved top-1 samples. The details are shown in Appendix~\ref{app:sim}.

Figure~\ref{fig:sim} illustrates that the evolution process does not yield higher similarity scores. Furthermore, similarity scores across all rounds remain relatively low. These findings indicate that the primary source of performance gain is the introduction of more complex data.
\section{Conclusion and Future Work}

This paper introduces \modelname{} models, the \cname{} fine-tuned Code LLMs. The experimental results demonstrate that \modelname{} models achieve SOTA performance surpassing all existing open-source Code LLMs on five widely recognized code generation benchmarks: HumanEval, HumanEval+, MBPP, DS-1000 and MultiPL-E. Notably, \modelname{} \textit{15B} model surpasses some of the well-known closed LLMs, such as Claude and Bard. Additionally, \modelname{} \textit{34B} achieves a HumanEval score comparable to GPT3.5 (ChatGPT) and surpasses it on the HumanEval+ benchmark. Furthermore, our analysis underscores the pivotal role of instruction complexity in enhancing performance. For future work, as depicted in Figure~\ref{fig:pass1}, our model still falls significantly behind the SOTA LLM, GPT4. Therefore, future work will further augment the performance of our model.


\section*{Acknowledgments}
This work is partially supported by National Natural Science Foundation of China Young Scientists Fund(No. 62206233) and Hong Kong RGC ECS (No. 22200722).

\bibliography{iclr2024_conference}

\begin{thebibliography}{51}
\providecommand{\natexlab}[1]{#1}
\providecommand{\url}[1]{\texttt{#1}}
\expandafter\ifx\csname urlstyle\endcsname\relax
  \providecommand{\doi}[1]{doi: #1}\else
  \providecommand{\doi}{doi: \begingroup \urlstyle{rm}\Url}\fi

\bibitem[Anil et~al.(2023)Anil, Dai, Firat, Johnson, Lepikhin, Passos, Shakeri, Taropa, Bailey, Chen, Chu, Clark, Shafey, Huang, Meier{-}Hellstern, Mishra, Moreira, Omernick, Robinson, Ruder, Tay, Xiao, Xu, Zhang, {\'{A}}brego, Ahn, Austin, Barham, Botha, Bradbury, Brahma, Brooks, Catasta, Cheng, Cherry, Choquette{-}Choo, Chowdhery, Crepy, Dave, Dehghani, Dev, Devlin, D{\'{\i}}az, Du, Dyer, Feinberg, Feng, Fienber, Freitag, Garcia, Gehrmann, Gonzalez, and et~al.]{palm2}
Rohan Anil, Andrew~M. Dai, Orhan Firat, Melvin Johnson, Dmitry Lepikhin, Alexandre Passos, Siamak Shakeri, Emanuel Taropa, Paige Bailey, Zhifeng Chen, Eric Chu, Jonathan~H. Clark, Laurent~El Shafey, Yanping Huang, Kathy Meier{-}Hellstern, Gaurav Mishra, Erica Moreira, Mark Omernick, Kevin Robinson, Sebastian Ruder, Yi~Tay, Kefan Xiao, Yuanzhong Xu, Yujing Zhang, Gustavo~Hern{\'{a}}ndez {\'{A}}brego, Junwhan Ahn, Jacob Austin, Paul Barham, Jan~A. Botha, James Bradbury, Siddhartha Brahma, Kevin Brooks, Michele Catasta, Yong Cheng, Colin Cherry, Christopher~A. Choquette{-}Choo, Aakanksha Chowdhery, Cl{\'{e}}ment Crepy, Shachi Dave, Mostafa Dehghani, Sunipa Dev, Jacob Devlin, Mark D{\'{\i}}az, Nan Du, Ethan Dyer, Vladimir Feinberg, Fangxiaoyu Feng, Vlad Fienber, Markus Freitag, Xavier Garcia, Sebastian Gehrmann, Lucas Gonzalez, and et~al.
\newblock Palm 2 technical report.
\newblock \emph{CoRR}, abs/2305.10403, 2023.
\newblock \doi{10.48550/arXiv.2305.10403}.
\newblock URL \url{https://doi.org/10.48550/arXiv.2305.10403}.

\bibitem[Aribandi et~al.(2022)Aribandi, Tay, Schuster, Rao, Zheng, Mehta, Zhuang, Tran, Bahri, Ni, Gupta, Hui, Ruder, and Metzler]{ExT5}
Vamsi Aribandi, Yi~Tay, Tal Schuster, Jinfeng Rao, Huaixiu~Steven Zheng, Sanket~Vaibhav Mehta, Honglei Zhuang, Vinh~Q. Tran, Dara Bahri, Jianmo Ni, Jai~Prakash Gupta, Kai Hui, Sebastian Ruder, and Donald Metzler.
\newblock Ext5: Towards extreme multi-task scaling for transfer learning.
\newblock In \emph{The Tenth International Conference on Learning Representations, {ICLR} 2022, Virtual Event, April 25-29, 2022}. OpenReview.net, 2022.
\newblock URL \url{https://openreview.net/forum?id=Vzh1BFUCiIX}.

\bibitem[Austin et~al.(2021)Austin, Odena, Nye, Bosma, Michalewski, Dohan, Jiang, Cai, Terry, Le, and Sutton]{MBPP}
Jacob Austin, Augustus Odena, Maxwell~I. Nye, Maarten Bosma, Henryk Michalewski, David Dohan, Ellen Jiang, Carrie~J. Cai, Michael Terry, Quoc~V. Le, and Charles Sutton.
\newblock Program synthesis with large language models.
\newblock \emph{CoRR}, abs/2108.07732, 2021.
\newblock URL \url{https://arxiv.org/abs/2108.07732}.

\bibitem[Black et~al.(2022)Black, Biderman, Hallahan, Anthony, Gao, Golding, He, Leahy, McDonell, Phang, Pieler, Prashanth, Purohit, Reynolds, Tow, Wang, and Weinbach]{GPT-NeoX-20B}
Sid Black, Stella Biderman, Eric Hallahan, Quentin Anthony, Leo Gao, Laurence Golding, Horace He, Connor Leahy, Kyle McDonell, Jason Phang, Michael Pieler, USVSN~Sai Prashanth, Shivanshu Purohit, Laria Reynolds, Jonathan Tow, Ben Wang, and Samuel Weinbach.
\newblock Gpt-neox-20b: An open-source autoregressive language model.
\newblock \emph{CoRR}, abs/2204.06745, 2022.
\newblock \doi{10.48550/arXiv.2204.06745}.
\newblock URL \url{https://doi.org/10.48550/arXiv.2204.06745}.

\bibitem[Brown et~al.(2020)Brown, Mann, Ryder, Subbiah, Kaplan, Dhariwal, Neelakantan, Shyam, Sastry, Askell, Agarwal, Herbert{-}Voss, Krueger, Henighan, Child, Ramesh, Ziegler, Wu, Winter, Hesse, Chen, Sigler, Litwin, Gray, Chess, Clark, Berner, McCandlish, Radford, Sutskever, and Amodei]{GPT3}
Tom~B. Brown, Benjamin Mann, Nick Ryder, Melanie Subbiah, Jared Kaplan, Prafulla Dhariwal, Arvind Neelakantan, Pranav Shyam, Girish Sastry, Amanda Askell, Sandhini Agarwal, Ariel Herbert{-}Voss, Gretchen Krueger, Tom Henighan, Rewon Child, Aditya Ramesh, Daniel~M. Ziegler, Jeffrey Wu, Clemens Winter, Christopher Hesse, Mark Chen, Eric Sigler, Mateusz Litwin, Scott Gray, Benjamin Chess, Jack Clark, Christopher Berner, Sam McCandlish, Alec Radford, Ilya Sutskever, and Dario Amodei.
\newblock Language models are few-shot learners.
\newblock In Hugo Larochelle, Marc'Aurelio Ranzato, Raia Hadsell, Maria{-}Florina Balcan, and Hsuan{-}Tien Lin (eds.), \emph{Advances in Neural Information Processing Systems 33: Annual Conference on Neural Information Processing Systems 2020, NeurIPS 2020, December 6-12, 2020, virtual}, 2020.
\newblock URL \url{https://proceedings.neurips.cc/paper/2020/hash/1457c0d6bfcb4967418bfb8ac142f64a-Abstract.html}.

\bibitem[Cassano et~al.(2022)Cassano, Gouwar, Nguyen, Nguyen, Phipps{-}Costin, Pinckney, Yee, Zi, Anderson, Feldman, Guha, Greenberg, and Jangda]{multipl_e}
Federico Cassano, John Gouwar, Daniel Nguyen, Sydney Nguyen, Luna Phipps{-}Costin, Donald Pinckney, Ming{-}Ho Yee, Yangtian Zi, Carolyn~Jane Anderson, Molly~Q. Feldman, Arjun Guha, Michael Greenberg, and Abhinav Jangda.
\newblock A scalable and extensible approach to benchmarking nl2code for 18 programming languages.
\newblock \emph{CoRR}, abs/2208.08227, 2022.
\newblock \doi{10.48550/arXiv.2208.08227}.
\newblock URL \url{https://doi.org/10.48550/arXiv.2208.08227}.

\bibitem[Chaudhary(2023)]{codealpaca}
Sahil Chaudhary.
\newblock Code alpaca: An instruction-following llama model for code generation.
\newblock \url{https://github.com/sahil280114/codealpaca}, 2023.

\bibitem[Chen et~al.(2021{\natexlab{a}})Chen, Tworek, Jun, Yuan, de~Oliveira~Pinto, Kaplan, Edwards, Burda, Joseph, Brockman, Ray, Puri, Krueger, Petrov, Khlaaf, Sastry, Mishkin, Chan, Gray, Ryder, Pavlov, Power, Kaiser, Bavarian, Winter, Tillet, Such, Cummings, Plappert, Chantzis, Barnes, Herbert{-}Voss, Guss, Nichol, Paino, Tezak, Tang, Babuschkin, Balaji, Jain, Saunders, Hesse, Carr, Leike, Achiam, Misra, Morikawa, Radford, Knight, Brundage, Murati, Mayer, Welinder, McGrew, Amodei, McCandlish, Sutskever, and Zaremba]{codex}
Mark Chen, Jerry Tworek, Heewoo Jun, Qiming Yuan, Henrique~Pond{\'{e}} de~Oliveira~Pinto, Jared Kaplan, Harrison Edwards, Yuri Burda, Nicholas Joseph, Greg Brockman, Alex Ray, Raul Puri, Gretchen Krueger, Michael Petrov, Heidy Khlaaf, Girish Sastry, Pamela Mishkin, Brooke Chan, Scott Gray, Nick Ryder, Mikhail Pavlov, Alethea Power, Lukasz Kaiser, Mohammad Bavarian, Clemens Winter, Philippe Tillet, Felipe~Petroski Such, Dave Cummings, Matthias Plappert, Fotios Chantzis, Elizabeth Barnes, Ariel Herbert{-}Voss, William~Hebgen Guss, Alex Nichol, Alex Paino, Nikolas Tezak, Jie Tang, Igor Babuschkin, Suchir Balaji, Shantanu Jain, William Saunders, Christopher Hesse, Andrew~N. Carr, Jan Leike, Joshua Achiam, Vedant Misra, Evan Morikawa, Alec Radford, Matthew Knight, Miles Brundage, Mira Murati, Katie Mayer, Peter Welinder, Bob McGrew, Dario Amodei, Sam McCandlish, Ilya Sutskever, and Wojciech Zaremba.
\newblock Evaluating large language models trained on code.
\newblock \emph{CoRR}, abs/2107.03374, 2021{\natexlab{a}}.
\newblock URL \url{https://arxiv.org/abs/2107.03374}.

\bibitem[Chen et~al.(2021{\natexlab{b}})Chen, Tworek, Jun, Yuan, de~Oliveira~Pinto, Kaplan, Edwards, Burda, Joseph, Brockman, Ray, Puri, Krueger, Petrov, Khlaaf, Sastry, Mishkin, Chan, Gray, Ryder, Pavlov, Power, Kaiser, Bavarian, Winter, Tillet, Such, Cummings, Plappert, Chantzis, Barnes, Herbert{-}Voss, Guss, Nichol, Paino, Tezak, Tang, Babuschkin, Balaji, Jain, Saunders, Hesse, Carr, Leike, Achiam, Misra, Morikawa, Radford, Knight, Brundage, Murati, Mayer, Welinder, McGrew, Amodei, McCandlish, Sutskever, and Zaremba]{humeval}
Mark Chen, Jerry Tworek, Heewoo Jun, Qiming Yuan, Henrique~Pond{\'{e}} de~Oliveira~Pinto, Jared Kaplan, Harrison Edwards, Yuri Burda, Nicholas Joseph, Greg Brockman, Alex Ray, Raul Puri, Gretchen Krueger, Michael Petrov, Heidy Khlaaf, Girish Sastry, Pamela Mishkin, Brooke Chan, Scott Gray, Nick Ryder, Mikhail Pavlov, Alethea Power, Lukasz Kaiser, Mohammad Bavarian, Clemens Winter, Philippe Tillet, Felipe~Petroski Such, Dave Cummings, Matthias Plappert, Fotios Chantzis, Elizabeth Barnes, Ariel Herbert{-}Voss, William~Hebgen Guss, Alex Nichol, Alex Paino, Nikolas Tezak, Jie Tang, Igor Babuschkin, Suchir Balaji, Shantanu Jain, William Saunders, Christopher Hesse, Andrew~N. Carr, Jan Leike, Joshua Achiam, Vedant Misra, Evan Morikawa, Alec Radford, Matthew Knight, Miles Brundage, Mira Murati, Katie Mayer, Peter Welinder, Bob McGrew, Dario Amodei, Sam McCandlish, Ilya Sutskever, and Wojciech Zaremba.
\newblock Evaluating large language models trained on code.
\newblock \emph{CoRR}, abs/2107.03374, 2021{\natexlab{b}}.
\newblock URL \url{https://arxiv.org/abs/2107.03374}.

\bibitem[Chiang et~al.(2023)Chiang, Li, Lin, Sheng, Wu, Zhang, Zheng, Zhuang, Zhuang, Gonzalez, Stoica, and Xing]{vicuna2023}
Wei-Lin Chiang, Zhuohan Li, Zi~Lin, Ying Sheng, Zhanghao Wu, Hao Zhang, Lianmin Zheng, Siyuan Zhuang, Yonghao Zhuang, Joseph~E. Gonzalez, Ion Stoica, and Eric~P. Xing.
\newblock Vicuna: An open-source chatbot impressing gpt-4 with 90\%* chatgpt quality, March 2023.
\newblock URL \url{https://vicuna.lmsys.org}.

\bibitem[Chowdhery et~al.(2022)Chowdhery, Narang, Devlin, Bosma, Mishra, Roberts, Barham, Chung, Sutton, Gehrmann, Schuh, Shi, Tsvyashchenko, Maynez, Rao, Barnes, Tay, Shazeer, Prabhakaran, Reif, Du, Hutchinson, Pope, Bradbury, Austin, Isard, Gur{-}Ari, Yin, Duke, Levskaya, Ghemawat, Dev, Michalewski, Garcia, Misra, Robinson, Fedus, Zhou, Ippolito, Luan, Lim, Zoph, Spiridonov, Sepassi, Dohan, Agrawal, Omernick, Dai, Pillai, Pellat, Lewkowycz, Moreira, Child, Polozov, Lee, Zhou, Wang, Saeta, Diaz, Firat, Catasta, Wei, Meier{-}Hellstern, Eck, Dean, Petrov, and Fiedel]{PaLM}
Aakanksha Chowdhery, Sharan Narang, Jacob Devlin, Maarten Bosma, Gaurav Mishra, Adam Roberts, Paul Barham, Hyung~Won Chung, Charles Sutton, Sebastian Gehrmann, Parker Schuh, Kensen Shi, Sasha Tsvyashchenko, Joshua Maynez, Abhishek Rao, Parker Barnes, Yi~Tay, Noam Shazeer, Vinodkumar Prabhakaran, Emily Reif, Nan Du, Ben Hutchinson, Reiner Pope, James Bradbury, Jacob Austin, Michael Isard, Guy Gur{-}Ari, Pengcheng Yin, Toju Duke, Anselm Levskaya, Sanjay Ghemawat, Sunipa Dev, Henryk Michalewski, Xavier Garcia, Vedant Misra, Kevin Robinson, Liam Fedus, Denny Zhou, Daphne Ippolito, David Luan, Hyeontaek Lim, Barret Zoph, Alexander Spiridonov, Ryan Sepassi, David Dohan, Shivani Agrawal, Mark Omernick, Andrew~M. Dai, Thanumalayan~Sankaranarayana Pillai, Marie Pellat, Aitor Lewkowycz, Erica Moreira, Rewon Child, Oleksandr Polozov, Katherine Lee, Zongwei Zhou, Xuezhi Wang, Brennan Saeta, Mark Diaz, Orhan Firat, Michele Catasta, Jason Wei, Kathy Meier{-}Hellstern, Douglas Eck, Jeff Dean, Slav Petrov, and Noah Fiedel.
\newblock Palm: Scaling language modeling with pathways.
\newblock \emph{CoRR}, abs/2204.02311, 2022.
\newblock \doi{10.48550/arXiv.2204.02311}.
\newblock URL \url{https://doi.org/10.48550/arXiv.2204.02311}.

\bibitem[Chung et~al.(2022)Chung, Hou, Longpre, Zoph, Tay, Fedus, Li, Wang, Dehghani, Brahma, Webson, Gu, Dai, Suzgun, Chen, Chowdhery, Narang, Mishra, Yu, Zhao, Huang, Dai, Yu, Petrov, Chi, Dean, Devlin, Roberts, Zhou, Le, and Wei]{flan-t5}
Hyung~Won Chung, Le~Hou, Shayne Longpre, Barret Zoph, Yi~Tay, William Fedus, Eric Li, Xuezhi Wang, Mostafa Dehghani, Siddhartha Brahma, Albert Webson, Shixiang~Shane Gu, Zhuyun Dai, Mirac Suzgun, Xinyun Chen, Aakanksha Chowdhery, Sharan Narang, Gaurav Mishra, Adams Yu, Vincent~Y. Zhao, Yanping Huang, Andrew~M. Dai, Hongkun Yu, Slav Petrov, Ed~H. Chi, Jeff Dean, Jacob Devlin, Adam Roberts, Denny Zhou, Quoc~V. Le, and Jason Wei.
\newblock Scaling instruction-finetuned language models.
\newblock \emph{CoRR}, abs/2210.11416, 2022.
\newblock \doi{10.48550/arXiv.2210.11416}.
\newblock URL \url{https://doi.org/10.48550/arXiv.2210.11416}.

\bibitem[Dettmers et~al.(2023)Dettmers, Pagnoni, Holtzman, and Zettlemoyer]{guanaco}
Tim Dettmers, Artidoro Pagnoni, Ari Holtzman, and Luke Zettlemoyer.
\newblock Qlora: Efficient finetuning of quantized llms.
\newblock \emph{CoRR}, abs/2305.14314, 2023.
\newblock \doi{10.48550/arXiv.2305.14314}.
\newblock URL \url{https://doi.org/10.48550/arXiv.2305.14314}.

\bibitem[Fried et~al.(2022)Fried, Aghajanyan, Lin, Wang, Wallace, Shi, Zhong, Yih, Zettlemoyer, and Lewis]{incoder}
Daniel Fried, Armen Aghajanyan, Jessy Lin, Sida Wang, Eric Wallace, Freda Shi, Ruiqi Zhong, Wen{-}tau Yih, Luke Zettlemoyer, and Mike Lewis.
\newblock Incoder: {A} generative model for code infilling and synthesis.
\newblock \emph{CoRR}, abs/2204.05999, 2022.
\newblock \doi{10.48550/arXiv.2204.05999}.
\newblock URL \url{https://doi.org/10.48550/arXiv.2204.05999}.

\bibitem[Hoffmann et~al.(2022)Hoffmann, Borgeaud, Mensch, Buchatskaya, Cai, Rutherford, de~Las~Casas, Hendricks, Welbl, Clark, Hennigan, Noland, Millican, van~den Driessche, Damoc, Guy, Osindero, Simonyan, Elsen, Rae, Vinyals, and Sifre]{Chinchilla}
Jordan Hoffmann, Sebastian Borgeaud, Arthur Mensch, Elena Buchatskaya, Trevor Cai, Eliza Rutherford, Diego de~Las~Casas, Lisa~Anne Hendricks, Johannes Welbl, Aidan Clark, Tom Hennigan, Eric Noland, Katie Millican, George van~den Driessche, Bogdan Damoc, Aurelia Guy, Simon Osindero, Karen Simonyan, Erich Elsen, Jack~W. Rae, Oriol Vinyals, and Laurent Sifre.
\newblock Training compute-optimal large language models.
\newblock \emph{CoRR}, abs/2203.15556, 2022.
\newblock \doi{10.48550/arXiv.2203.15556}.
\newblock URL \url{https://doi.org/10.48550/arXiv.2203.15556}.

\bibitem[Jha \& Reddy(2022)Jha and Reddy]{attackcode2}
Akshita Jha and Chandan~K. Reddy.
\newblock Codeattack: Code-based adversarial attacks for pre-trained programming language models.
\newblock In \emph{AAAI Conference on Artificial Intelligence}, 2022.
\newblock URL \url{https://api.semanticscholar.org/CorpusID:249240370}.

\bibitem[Khashabi et~al.(2020)Khashabi, Min, Khot, Sabharwal, Tafjord, Clark, and Hajishirzi]{UnifiedQA}
Daniel Khashabi, Sewon Min, Tushar Khot, Ashish Sabharwal, Oyvind Tafjord, Peter Clark, and Hannaneh Hajishirzi.
\newblock Unifiedqa: Crossing format boundaries with a single {QA} system.
\newblock In Trevor Cohn, Yulan He, and Yang Liu (eds.), \emph{Findings of the Association for Computational Linguistics: {EMNLP} 2020, Online Event, 16-20 November 2020}, volume {EMNLP} 2020 of \emph{Findings of {ACL}}, pp.\  1896--1907. Association for Computational Linguistics, 2020.
\newblock \doi{10.18653/v1/2020.findings-emnlp.171}.
\newblock URL \url{https://doi.org/10.18653/v1/2020.findings-emnlp.171}.

\bibitem[Lai et~al.(2022)Lai, Li, Wang, Zhang, Zhong, Zettlemoyer, Yih, Fried, Wang, and Yu]{DS1000}
Yuhang Lai, Chengxi Li, Yiming Wang, Tianyi Zhang, Ruiqi Zhong, Luke Zettlemoyer, Scott~Wen{-}tau Yih, Daniel Fried, Sida~I. Wang, and Tao Yu.
\newblock {DS-1000:} {A} natural and reliable benchmark for data science code generation.
\newblock \emph{CoRR}, abs/2211.11501, 2022.
\newblock \doi{10.48550/arXiv.2211.11501}.
\newblock URL \url{https://doi.org/10.48550/arXiv.2211.11501}.

\bibitem[Li et~al.(2023{\natexlab{a}})Li, Allal, Zi, Muennighoff, Kocetkov, Mou, Marone, Akiki, Li, Chim, et~al.]{li2023starcoder}
Raymond Li, Loubna~Ben Allal, Yangtian Zi, Niklas Muennighoff, Denis Kocetkov, Chenghao Mou, Marc Marone, Christopher Akiki, Jia Li, Jenny Chim, et~al.
\newblock Starcoder: may the source be with you!
\newblock \emph{arXiv preprint arXiv:2305.06161}, 2023{\natexlab{a}}.

\bibitem[Li et~al.(2022)Li, Choi, Chung, Kushman, Schrittwieser, Leblond, Eccles, Keeling, Gimeno, Lago, Hubert, Choy, de~Masson~d'Autume, Babuschkin, Chen, Huang, Welbl, Gowal, Cherepanov, Molloy, Mankowitz, Robson, Kohli, de~Freitas, Kavukcuoglu, and Vinyals]{AlphaCode}
Yujia Li, David~H. Choi, Junyoung Chung, Nate Kushman, Julian Schrittwieser, R{\'{e}}mi Leblond, Tom Eccles, James Keeling, Felix Gimeno, Agustin~Dal Lago, Thomas Hubert, Peter Choy, Cyprien de~Masson~d'Autume, Igor Babuschkin, Xinyun Chen, Po{-}Sen Huang, Johannes Welbl, Sven Gowal, Alexey Cherepanov, James Molloy, Daniel~J. Mankowitz, Esme~Sutherland Robson, Pushmeet Kohli, Nando de~Freitas, Koray Kavukcuoglu, and Oriol Vinyals.
\newblock Competition-level code generation with alphacode.
\newblock \emph{CoRR}, abs/2203.07814, 2022.
\newblock \doi{10.48550/arXiv.2203.07814}.
\newblock URL \url{https://doi.org/10.48550/arXiv.2203.07814}.

\bibitem[Li et~al.(2023{\natexlab{b}})Li, Zhang, Zhang, Long, Xie, and Zhang]{gte}
Zehan Li, Xin Zhang, Yanzhao Zhang, Dingkun Long, Pengjun Xie, and Meishan Zhang.
\newblock Towards general text embeddings with multi-stage contrastive learning.
\newblock \emph{CoRR}, abs/2308.03281, 2023{\natexlab{b}}.
\newblock \doi{10.48550/arXiv.2308.03281}.
\newblock URL \url{https://doi.org/10.48550/arXiv.2308.03281}.

\bibitem[Liu et~al.(2023)Liu, Xia, Wang, and Zhang]{humanevalp}
Jiawei Liu, Chunqiu~Steven Xia, Yuyao Wang, and Lingming Zhang.
\newblock Is your code generated by chatgpt really correct? rigorous evaluation of large language models for code generation.
\newblock \emph{CoRR}, abs/2305.01210, 2023.
\newblock \doi{10.48550/arXiv.2305.01210}.
\newblock URL \url{https://doi.org/10.48550/arXiv.2305.01210}.

\bibitem[Madaan et~al.(2023)Madaan, Shypula, Alon, Hashemi, Ranganathan, Yang, Neubig, and Yazdanbakhsh]{constraints}
Aman Madaan, Alex Shypula, Uri Alon, Milad Hashemi, Parthasarathy Ranganathan, Yiming Yang, Graham Neubig, and Amir Yazdanbakhsh.
\newblock Learning performance-improving code edits.
\newblock \emph{ArXiv}, abs/2302.07867, 2023.
\newblock URL \url{https://api.semanticscholar.org/CorpusID:256868633}.

\bibitem[Microsoft(2023)]{Azure}
Microsoft.
\newblock Azure openai service models.
\newblock \url{https://learn.microsoft.com/en-us/azure/cognitive-services/openai/concepts/models}, 2023.

\bibitem[Muennighoff et~al.(2023)Muennighoff, Liu, Zebaze, Zheng, Hui, Zhuo, Singh, Tang, von Werra, and Longpre]{octocoder}
Niklas Muennighoff, Qian Liu, Armel Zebaze, Qinkai Zheng, Binyuan Hui, Terry~Yue Zhuo, Swayam Singh, Xiangru Tang, Leandro von Werra, and Shayne Longpre.
\newblock Octopack: Instruction tuning code large language models.
\newblock \emph{CoRR}, abs/2308.07124, 2023.
\newblock \doi{10.48550/arXiv.2308.07124}.
\newblock URL \url{https://doi.org/10.48550/arXiv.2308.07124}.

\bibitem[Nijkamp et~al.(2023{\natexlab{a}})Nijkamp, Hayashi, Xiong, Savarese, and Zhou]{codegen2}
Erik Nijkamp, Hiroaki Hayashi, Caiming Xiong, Silvio Savarese, and Yingbo Zhou.
\newblock Codegen2: Lessons for training llms on programming and natural languages.
\newblock \emph{CoRR}, abs/2305.02309, 2023{\natexlab{a}}.
\newblock \doi{10.48550/arXiv.2305.02309}.
\newblock URL \url{https://doi.org/10.48550/arXiv.2305.02309}.

\bibitem[Nijkamp et~al.(2023{\natexlab{b}})Nijkamp, Pang, Hayashi, Tu, Wang, Zhou, Savarese, and Xiong]{codegen}
Erik Nijkamp, Bo~Pang, Hiroaki Hayashi, Lifu Tu, Huan Wang, Yingbo Zhou, Silvio Savarese, and Caiming Xiong.
\newblock Codegen: An open large language model for code with multi-turn program synthesis.
\newblock In \emph{The Eleventh International Conference on Learning Representations}, 2023{\natexlab{b}}.
\newblock URL \url{https://openreview.net/forum?id=iaYcJKpY2B_}.

\bibitem[OpenAI(2023)]{GPT4}
OpenAI.
\newblock {GPT-4} technical report.
\newblock \emph{CoRR}, abs/2303.08774, 2023.
\newblock \doi{10.48550/arXiv.2303.08774}.
\newblock URL \url{https://doi.org/10.48550/arXiv.2303.08774}.

\bibitem[Ouyang et~al.(2022)Ouyang, Wu, Jiang, Almeida, Wainwright, Mishkin, Zhang, Agarwal, Slama, Ray, Schulman, Hilton, Kelton, Miller, Simens, Askell, Welinder, Christiano, Leike, and Lowe]{DBLP:conf/nips/Ouyang0JAWMZASR22}
Long Ouyang, Jeffrey Wu, Xu~Jiang, Diogo Almeida, Carroll~L. Wainwright, Pamela Mishkin, Chong Zhang, Sandhini Agarwal, Katarina Slama, Alex Ray, John Schulman, Jacob Hilton, Fraser Kelton, Luke Miller, Maddie Simens, Amanda Askell, Peter Welinder, Paul~F. Christiano, Jan Leike, and Ryan Lowe.
\newblock Training language models to follow instructions with human feedback.
\newblock In \emph{NeurIPS}, 2022.
\newblock URL \url{http://papers.nips.cc/paper\_files/paper/2022/hash/b1efde53be364a73914f58805a001731-Abstract-Conference.html}.

\bibitem[Penedo et~al.(2023)Penedo, Malartic, Hesslow, Cojocaru, Cappelli, Alobeidli, Pannier, Almazrouei, and Launay]{falcon}
Guilherme Penedo, Quentin Malartic, Daniel Hesslow, Ruxandra Cojocaru, Alessandro Cappelli, Hamza Alobeidli, Baptiste Pannier, Ebtesam Almazrouei, and Julien Launay.
\newblock The refinedweb dataset for falcon {LLM:} outperforming curated corpora with web data, and web data only.
\newblock \emph{CoRR}, abs/2306.01116, 2023.
\newblock \doi{10.48550/arXiv.2306.01116}.
\newblock URL \url{https://doi.org/10.48550/arXiv.2306.01116}.

\bibitem[Rae et~al.(2021)Rae, Borgeaud, Cai, Millican, Hoffmann, Song, Aslanides, Henderson, Ring, Young, Rutherford, Hennigan, Menick, Cassirer, Powell, van~den Driessche, Hendricks, Rauh, Huang, Glaese, Welbl, Dathathri, Huang, Uesato, Mellor, Higgins, Creswell, McAleese, Wu, Elsen, Jayakumar, Buchatskaya, Budden, Sutherland, Simonyan, Paganini, Sifre, Martens, Li, Kuncoro, Nematzadeh, Gribovskaya, Donato, Lazaridou, Mensch, Lespiau, Tsimpoukelli, Grigorev, Fritz, Sottiaux, Pajarskas, Pohlen, Gong, Toyama, de~Masson~d'Autume, Li, Terzi, Mikulik, Babuschkin, Clark, de~Las~Casas, Guy, Jones, Bradbury, Johnson, Hechtman, Weidinger, Gabriel, Isaac, Lockhart, Osindero, Rimell, Dyer, Vinyals, Ayoub, Stanway, Bennett, Hassabis, Kavukcuoglu, and Irving]{gopher}
Jack~W. Rae, Sebastian Borgeaud, Trevor Cai, Katie Millican, Jordan Hoffmann, H.~Francis Song, John Aslanides, Sarah Henderson, Roman Ring, Susannah Young, Eliza Rutherford, Tom Hennigan, Jacob Menick, Albin Cassirer, Richard Powell, George van~den Driessche, Lisa~Anne Hendricks, Maribeth Rauh, Po{-}Sen Huang, Amelia Glaese, Johannes Welbl, Sumanth Dathathri, Saffron Huang, Jonathan Uesato, John Mellor, Irina Higgins, Antonia Creswell, Nat McAleese, Amy Wu, Erich Elsen, Siddhant~M. Jayakumar, Elena Buchatskaya, David Budden, Esme Sutherland, Karen Simonyan, Michela Paganini, Laurent Sifre, Lena Martens, Xiang~Lorraine Li, Adhiguna Kuncoro, Aida Nematzadeh, Elena Gribovskaya, Domenic Donato, Angeliki Lazaridou, Arthur Mensch, Jean{-}Baptiste Lespiau, Maria Tsimpoukelli, Nikolai Grigorev, Doug Fritz, Thibault Sottiaux, Mantas Pajarskas, Toby Pohlen, Zhitao Gong, Daniel Toyama, Cyprien de~Masson~d'Autume, Yujia Li, Tayfun Terzi, Vladimir Mikulik, Igor Babuschkin, Aidan Clark, Diego de~Las~Casas, Aurelia Guy,
  Chris Jones, James Bradbury, Matthew~J. Johnson, Blake~A. Hechtman, Laura Weidinger, Iason Gabriel, William Isaac, Edward Lockhart, Simon Osindero, Laura Rimell, Chris Dyer, Oriol Vinyals, Kareem Ayoub, Jeff Stanway, Lorrayne Bennett, Demis Hassabis, Koray Kavukcuoglu, and Geoffrey Irving.
\newblock Scaling language models: Methods, analysis {\&} insights from training gopher.
\newblock \emph{CoRR}, abs/2112.11446, 2021.
\newblock URL \url{https://arxiv.org/abs/2112.11446}.

\bibitem[Raffel et~al.(2020)Raffel, Shazeer, Roberts, Lee, Narang, Matena, Zhou, Li, and Liu]{t5}
Colin Raffel, Noam Shazeer, Adam Roberts, Katherine Lee, Sharan Narang, Michael Matena, Yanqi Zhou, Wei Li, and Peter~J. Liu.
\newblock Exploring the limits of transfer learning with a unified text-to-text transformer.
\newblock \emph{J. Mach. Learn. Res.}, 21:\penalty0 140:1--140:67, 2020.
\newblock URL \url{http://jmlr.org/papers/v21/20-074.html}.

\bibitem[Rozi{\`{e}}re et~al.(2023)Rozi{\`{e}}re, Gehring, Gloeckle, Sootla, Gat, Tan, Adi, Liu, Remez, Rapin, Kozhevnikov, Evtimov, Bitton, Bhatt, Canton{-}Ferrer, Grattafiori, Xiong, D{\'{e}}fossez, Copet, Azhar, Touvron, Martin, Usunier, Scialom, and Synnaeve]{codellama}
Baptiste Rozi{\`{e}}re, Jonas Gehring, Fabian Gloeckle, Sten Sootla, Itai Gat, Xiaoqing~Ellen Tan, Yossi Adi, Jingyu Liu, Tal Remez, J{\'{e}}r{\'{e}}my Rapin, Artyom Kozhevnikov, Ivan Evtimov, Joanna Bitton, Manish Bhatt, Cristian Canton{-}Ferrer, Aaron Grattafiori, Wenhan Xiong, Alexandre D{\'{e}}fossez, Jade Copet, Faisal Azhar, Hugo Touvron, Louis Martin, Nicolas Usunier, Thomas Scialom, and Gabriel Synnaeve.
\newblock Code llama: Open foundation models for code.
\newblock \emph{CoRR}, abs/2308.12950, 2023.
\newblock \doi{10.48550/arXiv.2308.12950}.
\newblock URL \url{https://doi.org/10.48550/arXiv.2308.12950}.

\bibitem[Sanh et~al.(2022)Sanh, Webson, Raffel, Bach, Sutawika, Alyafeai, Chaffin, Stiegler, Raja, Dey, Bari, Xu, Thakker, Sharma, Szczechla, Kim, Chhablani, Nayak, Datta, Chang, Jiang, Wang, Manica, Shen, Yong, Pandey, Bawden, Wang, Neeraj, Rozen, Sharma, Santilli, F{\'{e}}vry, Fries, Teehan, Scao, Biderman, Gao, Wolf, and Rush]{T0}
Victor Sanh, Albert Webson, Colin Raffel, Stephen~H. Bach, Lintang Sutawika, Zaid Alyafeai, Antoine Chaffin, Arnaud Stiegler, Arun Raja, Manan Dey, M~Saiful Bari, Canwen Xu, Urmish Thakker, Shanya~Sharma Sharma, Eliza Szczechla, Taewoon Kim, Gunjan Chhablani, Nihal~V. Nayak, Debajyoti Datta, Jonathan Chang, Mike~Tian{-}Jian Jiang, Han Wang, Matteo Manica, Sheng Shen, Zheng~Xin Yong, Harshit Pandey, Rachel Bawden, Thomas Wang, Trishala Neeraj, Jos Rozen, Abheesht Sharma, Andrea Santilli, Thibault F{\'{e}}vry, Jason~Alan Fries, Ryan Teehan, Teven~Le Scao, Stella Biderman, Leo Gao, Thomas Wolf, and Alexander~M. Rush.
\newblock Multitask prompted training enables zero-shot task generalization.
\newblock In \emph{The Tenth International Conference on Learning Representations, {ICLR} 2022, Virtual Event, April 25-29, 2022}. OpenReview.net, 2022.
\newblock URL \url{https://openreview.net/forum?id=9Vrb9D0WI4}.

\bibitem[Shen et~al.(2023)Shen, Zhang, Chen, Zan, Geng, Fu, Zeng, Yu, Ji, Zhao, Guo, and Wang]{pangucoder2}
Bo~Shen, Jiaxin Zhang, Taihong Chen, Daoguang Zan, Bing Geng, An~Fu, Muhan Zeng, Ailun Yu, Jichuan Ji, Jingyang Zhao, Yuenan Guo, and Qianxiang Wang.
\newblock Pangu-coder2: Boosting large language models for code with ranking feedback.
\newblock \emph{CoRR}, abs/2307.14936, 2023.
\newblock \doi{10.48550/arXiv.2307.14936}.
\newblock URL \url{https://doi.org/10.48550/arXiv.2307.14936}.

\bibitem[Taori et~al.(2023)Taori, Gulrajani, Zhang, Dubois, Li, Guestrin, Liang, and Hashimoto]{alpaca}
Rohan Taori, Ishaan Gulrajani, Tianyi Zhang, Yann Dubois, Xuechen Li, Carlos Guestrin, Percy Liang, and Tatsunori~B. Hashimoto.
\newblock Stanford alpaca: An instruction-following llama model.
\newblock \url{https://github.com/tatsu-lab/stanford_alpaca}, 2023.

\bibitem[Tay et~al.(2022)Tay, Dehghani, Tran, Garcia, Bahri, Schuster, Zheng, Houlsby, and Metzler]{UL2}
Yi~Tay, Mostafa Dehghani, Vinh~Q. Tran, Xavier Garcia, Dara Bahri, Tal Schuster, Huaixiu~Steven Zheng, Neil Houlsby, and Donald Metzler.
\newblock Unifying language learning paradigms.
\newblock \emph{CoRR}, abs/2205.05131, 2022.
\newblock \doi{10.48550/arXiv.2205.05131}.
\newblock URL \url{https://doi.org/10.48550/arXiv.2205.05131}.

\bibitem[Thoppilan et~al.(2022)Thoppilan, Freitas, Hall, Shazeer, Kulshreshtha, Cheng, Jin, Bos, Baker, Du, Li, Lee, Zheng, Ghafouri, Menegali, Huang, Krikun, Lepikhin, Qin, Chen, Xu, Chen, Roberts, Bosma, Zhou, Chang, Krivokon, Rusch, Pickett, Meier{-}Hellstern, Morris, Doshi, Santos, Duke, Soraker, Zevenbergen, Prabhakaran, Diaz, Hutchinson, Olson, Molina, Hoffman{-}John, Lee, Aroyo, Rajakumar, Butryna, Lamm, Kuzmina, Fenton, Cohen, Bernstein, Kurzweil, Aguera{-}Arcas, Cui, Croak, Chi, and Le]{LaMDA}
Romal Thoppilan, Daniel~De Freitas, Jamie Hall, Noam Shazeer, Apoorv Kulshreshtha, Heng{-}Tze Cheng, Alicia Jin, Taylor Bos, Leslie Baker, Yu~Du, YaGuang Li, Hongrae Lee, Huaixiu~Steven Zheng, Amin Ghafouri, Marcelo Menegali, Yanping Huang, Maxim Krikun, Dmitry Lepikhin, James Qin, Dehao Chen, Yuanzhong Xu, Zhifeng Chen, Adam Roberts, Maarten Bosma, Yanqi Zhou, Chung{-}Ching Chang, Igor Krivokon, Will Rusch, Marc Pickett, Kathleen~S. Meier{-}Hellstern, Meredith~Ringel Morris, Tulsee Doshi, Renelito~Delos Santos, Toju Duke, Johnny Soraker, Ben Zevenbergen, Vinodkumar Prabhakaran, Mark Diaz, Ben Hutchinson, Kristen Olson, Alejandra Molina, Erin Hoffman{-}John, Josh Lee, Lora Aroyo, Ravi Rajakumar, Alena Butryna, Matthew Lamm, Viktoriya Kuzmina, Joe Fenton, Aaron Cohen, Rachel Bernstein, Ray Kurzweil, Blaise Aguera{-}Arcas, Claire Cui, Marian Croak, Ed~H. Chi, and Quoc Le.
\newblock Lamda: Language models for dialog applications.
\newblock \emph{CoRR}, abs/2201.08239, 2022.
\newblock URL \url{https://arxiv.org/abs/2201.08239}.

\bibitem[Touvron et~al.(2023{\natexlab{a}})Touvron, Lavril, Izacard, Martinet, Lachaux, Lacroix, Rozi{\`{e}}re, Goyal, Hambro, Azhar, Rodriguez, Joulin, Grave, and Lample]{llama}
Hugo Touvron, Thibaut Lavril, Gautier Izacard, Xavier Martinet, Marie{-}Anne Lachaux, Timoth{\'{e}}e Lacroix, Baptiste Rozi{\`{e}}re, Naman Goyal, Eric Hambro, Faisal Azhar, Aur{\'{e}}lien Rodriguez, Armand Joulin, Edouard Grave, and Guillaume Lample.
\newblock Llama: Open and efficient foundation language models.
\newblock \emph{CoRR}, abs/2302.13971, 2023{\natexlab{a}}.
\newblock \doi{10.48550/arXiv.2302.13971}.
\newblock URL \url{https://doi.org/10.48550/arXiv.2302.13971}.

\bibitem[Touvron et~al.(2023{\natexlab{b}})Touvron, Martin, Stone, Albert, Almahairi, Babaei, Bashlykov, Batra, Bhargava, Bhosale, Bikel, Blecher, Canton{-}Ferrer, Chen, Cucurull, Esiobu, Fernandes, Fu, Fu, Fuller, Gao, Goswami, Goyal, Hartshorn, Hosseini, Hou, Inan, Kardas, Kerkez, Khabsa, Kloumann, Korenev, Koura, Lachaux, Lavril, Lee, Liskovich, Lu, Mao, Martinet, Mihaylov, Mishra, Molybog, Nie, Poulton, Reizenstein, Rungta, Saladi, Schelten, Silva, Smith, Subramanian, Tan, Tang, Taylor, Williams, Kuan, Xu, Yan, Zarov, Zhang, Fan, Kambadur, Narang, Rodriguez, Stojnic, Edunov, and Scialom]{llama2}
Hugo Touvron, Louis Martin, Kevin Stone, Peter Albert, Amjad Almahairi, Yasmine Babaei, Nikolay Bashlykov, Soumya Batra, Prajjwal Bhargava, Shruti Bhosale, Dan Bikel, Lukas Blecher, Cristian Canton{-}Ferrer, Moya Chen, Guillem Cucurull, David Esiobu, Jude Fernandes, Jeremy Fu, Wenyin Fu, Brian Fuller, Cynthia Gao, Vedanuj Goswami, Naman Goyal, Anthony Hartshorn, Saghar Hosseini, Rui Hou, Hakan Inan, Marcin Kardas, Viktor Kerkez, Madian Khabsa, Isabel Kloumann, Artem Korenev, Punit~Singh Koura, Marie{-}Anne Lachaux, Thibaut Lavril, Jenya Lee, Diana Liskovich, Yinghai Lu, Yuning Mao, Xavier Martinet, Todor Mihaylov, Pushkar Mishra, Igor Molybog, Yixin Nie, Andrew Poulton, Jeremy Reizenstein, Rashi Rungta, Kalyan Saladi, Alan Schelten, Ruan Silva, Eric~Michael Smith, Ranjan Subramanian, Xiaoqing~Ellen Tan, Binh Tang, Ross Taylor, Adina Williams, Jian~Xiang Kuan, Puxin Xu, Zheng Yan, Iliyan Zarov, Yuchen Zhang, Angela Fan, Melanie Kambadur, Sharan Narang, Aur{\'{e}}lien Rodriguez, Robert Stojnic, Sergey Edunov,
  and Thomas Scialom.
\newblock Llama 2: Open foundation and fine-tuned chat models.
\newblock \emph{CoRR}, abs/2307.09288, 2023{\natexlab{b}}.
\newblock \doi{10.48550/arXiv.2307.09288}.
\newblock URL \url{https://doi.org/10.48550/arXiv.2307.09288}.

\bibitem[Wang \& Komatsuzaki(2021)Wang and Komatsuzaki]{gpt-j}
Ben Wang and Aran Komatsuzaki.
\newblock {GPT-J-6B: A 6 Billion Parameter Autoregressive Language Model}.
\newblock \url{https://github.com/kingoflolz/mesh-transformer-jax}, May 2021.

\bibitem[Wang et~al.(2022)Wang, Kordi, Mishra, Liu, Smith, Khashabi, and Hajishirzi]{wang2022self}
Yizhong Wang, Yeganeh Kordi, Swaroop Mishra, Alisa Liu, Noah~A Smith, Daniel Khashabi, and Hannaneh Hajishirzi.
\newblock Self-instruct: Aligning language model with self generated instructions.
\newblock \emph{arXiv preprint arXiv:2212.10560}, 2022.

\bibitem[Wang et~al.(2021)Wang, Wang, Joty, and Hoi]{codet5}
Yue Wang, Weishi Wang, Shafiq~R. Joty, and Steven C.~H. Hoi.
\newblock Codet5: Identifier-aware unified pre-trained encoder-decoder models for code understanding and generation.
\newblock In Marie{-}Francine Moens, Xuanjing Huang, Lucia Specia, and Scott~Wen{-}tau Yih (eds.), \emph{Proceedings of the 2021 Conference on Empirical Methods in Natural Language Processing, {EMNLP} 2021, Virtual Event / Punta Cana, Dominican Republic, 7-11 November, 2021}, pp.\  8696--8708. Association for Computational Linguistics, 2021.
\newblock \doi{10.18653/v1/2021.emnlp-main.685}.
\newblock URL \url{https://doi.org/10.18653/v1/2021.emnlp-main.685}.

\bibitem[Wang et~al.(2023)Wang, Le, Gotmare, Bui, Li, and Hoi]{CodeT5+}
Yue Wang, Hung Le, Akhilesh~Deepak Gotmare, Nghi D.~Q. Bui, Junnan Li, and Steven C.~H. Hoi.
\newblock Codet5+: Open code large language models for code understanding and generation.
\newblock \emph{CoRR}, abs/2305.07922, 2023.
\newblock \doi{10.48550/arXiv.2305.07922}.
\newblock URL \url{https://doi.org/10.48550/arXiv.2305.07922}.

\bibitem[Wei et~al.(2022)Wei, Bosma, Zhao, Guu, Yu, Lester, Du, Dai, and Le]{DBLP:conf/iclr/WeiBZGYLDDL22}
Jason Wei, Maarten Bosma, Vincent~Y. Zhao, Kelvin Guu, Adams~Wei Yu, Brian Lester, Nan Du, Andrew~M. Dai, and Quoc~V. Le.
\newblock Finetuned language models are zero-shot learners.
\newblock In \emph{The Tenth International Conference on Learning Representations, {ICLR} 2022, Virtual Event, April 25-29, 2022}. OpenReview.net, 2022.
\newblock URL \url{https://openreview.net/forum?id=gEZrGCozdqR}.

\bibitem[Xu et~al.(2023)Xu, Sun, Zheng, Geng, Zhao, Feng, Tao, and Jiang]{xu2023wizardlm}
Can Xu, Qingfeng Sun, Kai Zheng, Xiubo Geng, Pu~Zhao, Jiazhan Feng, Chongyang Tao, and Daxin Jiang.
\newblock Wizardlm: Empowering large language models to follow complex instructions.
\newblock \emph{arXiv preprint arXiv:2304.12244}, 2023.

\bibitem[Xu et~al.(2022)Xu, Chen, Du, Shao, Wang, Li, and Yang]{ZeroPrompt}
Hanwei Xu, Yujun Chen, Yulun Du, Nan Shao, Yanggang Wang, Haiyu Li, and Zhilin Yang.
\newblock Zeroprompt: Scaling prompt-based pretraining to 1, 000 tasks improves zero-shot generalization.
\newblock In Yoav Goldberg, Zornitsa Kozareva, and Yue Zhang (eds.), \emph{Findings of the Association for Computational Linguistics: {EMNLP} 2022, Abu Dhabi, United Arab Emirates, December 7-11, 2022}, pp.\  4235--4252. Association for Computational Linguistics, 2022.
\newblock URL \url{https://aclanthology.org/2022.findings-emnlp.312}.

\bibitem[Yang et~al.(2022)Yang, Shi, He, and Lo]{attackcode}
Zhou Yang, Jieke Shi, Junda He, and David Lo.
\newblock Natural attack for pre-trained models of code.
\newblock \emph{2022 IEEE/ACM 44th International Conference on Software Engineering (ICSE)}, pp.\  1482--1493, 2022.
\newblock URL \url{https://api.semanticscholar.org/CorpusID:246210250}.

\bibitem[Zeng et~al.(2022)Zeng, Liu, Du, Wang, Lai, Ding, Yang, Xu, Zheng, Xia, Tam, Ma, Xue, Zhai, Chen, Zhang, Dong, and Tang]{GLM-130B}
Aohan Zeng, Xiao Liu, Zhengxiao Du, Zihan Wang, Hanyu Lai, Ming Ding, Zhuoyi Yang, Yifan Xu, Wendi Zheng, Xiao Xia, Weng~Lam Tam, Zixuan Ma, Yufei Xue, Jidong Zhai, Wenguang Chen, Peng Zhang, Yuxiao Dong, and Jie Tang.
\newblock {GLM-130B:} an open bilingual pre-trained model.
\newblock \emph{CoRR}, abs/2210.02414, 2022.
\newblock \doi{10.48550/arXiv.2210.02414}.
\newblock URL \url{https://doi.org/10.48550/arXiv.2210.02414}.

\bibitem[Zhang et~al.(2022)Zhang, Roller, Goyal, Artetxe, Chen, Chen, Dewan, Diab, Li, Lin, Mihaylov, Ott, Shleifer, Shuster, Simig, Koura, Sridhar, Wang, and Zettlemoyer]{opt}
Susan Zhang, Stephen Roller, Naman Goyal, Mikel Artetxe, Moya Chen, Shuohui Chen, Christopher Dewan, Mona~T. Diab, Xian Li, Xi~Victoria Lin, Todor Mihaylov, Myle Ott, Sam Shleifer, Kurt Shuster, Daniel Simig, Punit~Singh Koura, Anjali Sridhar, Tianlu Wang, and Luke Zettlemoyer.
\newblock {OPT:} open pre-trained transformer language models.
\newblock \emph{CoRR}, abs/2205.01068, 2022.
\newblock \doi{10.48550/arXiv.2205.01068}.
\newblock URL \url{https://doi.org/10.48550/arXiv.2205.01068}.

\bibitem[Zheng et~al.(2023)Zheng, Xia, Zou, Dong, Wang, Xue, Wang, Shen, Wang, Li, Su, Yang, and Tang]{CodeGeeX}
Qinkai Zheng, Xiao Xia, Xu~Zou, Yuxiao Dong, Shan Wang, Yufei Xue, Zihan Wang, Lei Shen, Andi Wang, Yang Li, Teng Su, Zhilin Yang, and Jie Tang.
\newblock Codegeex: {A} pre-trained model for code generation with multilingual evaluations on humaneval-x.
\newblock \emph{CoRR}, abs/2303.17568, 2023.
\newblock \doi{10.48550/arXiv.2303.17568}.
\newblock URL \url{https://doi.org/10.48550/arXiv.2303.17568}.

\end{thebibliography}
\bibliographystyle{iclr2024_conference}

\appendix
\newpage
\section{Prompt Formats}\label{app:prompt}

In this section, we include the prompt for evaluation on different tasks.

\newenvironment{myblock4}{%
  \begin{tcolorbox}[colback=beaublue!8!white,colframe=beaublue!10!black,title=Zero-Shot Prompt for Evaluation on HumanEval and HumanEval+]
}{%
  \end{tcolorbox}
}
\begin{myblock4}
Below is an instruction that describes a task, paired with an input that provides further context. Write a response that appropriately completes the request. \\\\\#\#\# Instruction:\\
Create a Python script for this problem:\\\{Question\}\\ \\\#\#\# Response:
\end{myblock4}

\newenvironment{myblock5}{%
  \begin{tcolorbox}[colback=beaublue!8!white,colframe=beaublue!10!black,title=Three-Shot Prompt for Evaluation on MBPP]
}{%
  \end{tcolorbox}
}
\begin{myblock5}
Below is an instruction that describes a task, paired with an input that provides further context. Write a response that appropriately completes the request. \\\\\#\#\# Instruction:\\
Create a Python script for this problem:\\\{Question\}\\\{Test Example 1\}\\\{Test Example 2\}\\\{Test Example 3\}\\\\\#\#\# Response:
\end{myblock5}

\newenvironment{myblock7}{%
  \begin{tcolorbox}[colback=beaublue!8!white,colframe=beaublue!10!black,title=Zero-Shot Prompt for Evaluation on DS-1000 (Completion)]
}{%
  \end{tcolorbox}
}
\begin{myblock7}
Below is an instruction that describes a task, paired with an input that provides further context. Write a response that appropriately completes the request. \\\\\#\#\# Instruction:\\
\{Question\}\\Complete the Python code in "...".\\\\\#\#\# Response:
\end{myblock7}

In the case of DS-1000 (Insertion), adherence to the benchmark's specifications necessitates the utilization of StarCoder's specialized insertion symbol. Consequently, we have found it imperative to align with the same prompt format employed by StarCoder for this particular benchmark.

For the MultiPL-E benchmark, we recognized the need to align with the evaluation codes provided by bigcode-evaluation-harness.\footnote{\url{https://github.com/bigcode-project/bigcode-evaluation-harness}} Consequently, we opted to adopt the same prompt format utilized by StarCoder. 



\section{Baselines Details}\label{app:baselines}

We include a large amount of models as our baselines. For GPT3.5 (ChatGPT)\&GPT4. their results are obtained from GPT4's report and EvalPlus. The results of Code-Davinci-002, Code-Cushman-001, Codex, PaLM, PaLM 2, LaMDA, AlpahaCode, Incoder, StarCoder, LLaMa, CodeGen, CodeGeeX, CodeT5+, and InstructCodeT5+ are from StarCoder or CodeT5+'s paper. The results of Bard are evaluated with Google's API. The results of Claude are evaluated with Anthropic's API. The results of Instruct-Codegen-16B, Guanaco-65B, Falcon-40B-Instruct, and Vicuna-13B are evaluated with the open-sourced checkpoints. The results of CodeLlama-Series are from CodeLlama's paper. The results of OctoCoder are from its paper. The results of PanguCoder2 are also from its paper.

The MBPP score of StarCoder differs from that in its original paper. Through a personal contact, we were informed that StarCoder was evaluated using a cleaned and smaller version of MBPP, comprising only 397 problems, significantly fewer than the original MBPP benchmarks (500). Consequently, we conducted a re-evaluation of StarCoder using the original MBPP.

\section{Similarity Checking and Data Filtering}\label{app:sim}

The prompt formats to compute the similarity score are as follow:
\newenvironment{myblock8}{%
  \begin{tcolorbox}[colback=beaublue!8!white,colframe=beaublue!10!black,title=System Prompt for Similarity Checking]
}{%
  \end{tcolorbox}
}
\begin{myblock8}
Your task is to evaluate the similarity of the two given coding tasks. Please review the two coding tasks carefully, paying close attention to the overlap in function names, code structures, topics, and contents. Once you have carefully reviewed both coding tasks, provide a similarity score between these two coding tasks. The score should range from 1 to 10 (1: completely different coding tasks; 10: identical coding tasks). You only need to provide your score. The response format is:\\
Score: '...'
\end{myblock8}

\newenvironment{myblock9}{%
  \begin{tcolorbox}[colback=beaublue!8!white,colframe=beaublue!10!black,title=User Input for Similarity Checking]
}{%
  \end{tcolorbox}
}
\begin{myblock9}
\# Task1\\
\{task1\}
\\\\
\# Task2\\
\{task2\}
\end{myblock9}

To thoroughly prevent data leakage from the test datasets to the training dataset, we implemented an additional data filtering step. Utilizing the SOTA embeddings model, gte-large, we treated all test samples as queries to extract the top 5 samples from the training data. Following this, GPT-4 was employed to evaluate the similarity between the retrieved samples and the test sample. The task for GPT-4 is simplified to a binary decision—either a ``yes" or ``no" indicating a match. In case of a positive match, the sample is excluded from the training data.

\section{Evol Examples}\label{app:evol_example}

In this section, we present some evolved examples to elucidate the influence exerted by our \cname{}.

Example 1:
\begin{itemize}
    \item Round 0: Write a MongoDB query to select all documents in a collection where the field 'category' is 'clothes'.
    \item Round 1: Write a MongoDB query to select all documents in a collection where the field 'category' is 'clothes' and the 'brand' field is not equal to 'Nike'.
    \item Round 2: Write a MongoDB query to select all documents in a collection where the field 'category' is 'clothes' and the 'brand' field is not equal to 'Nike', and the 'price' field is greater than or equal to 100 and less than or equal to 500.
    \item Round 3: Write a MongoDB query to select all documents in a collection where the field 'category' is 'clothes' and the 'brand' field is not equal to 'Nike', and the 'price' field is greater than or equal to 100 and less than or equal to 500, and the 'color' field is either 'red' or 'blue'. Additionally, sort the documents in descending order by the 'date\_added' field and limit the result to the first 10 documents.
\end{itemize}

Example 2:
\begin{itemize}
    \item Round 0: Using Tensorflow, set up a model to predict the classification of a handwritten digit.
    \item Round 1: Using Tensorflow, set up a model to predict the classification of a handwritten digit with the constraint that the model must be trained using only 1000 samples from the MNIST dataset.
    \item Round 2: Using TensorFlow, set up a model to predict the classification of a handwritten digit with the constraint that the model must be trained using only 500 samples from the MNIST dataset. Additionally, implement transfer learning to incorporate the features learned from a pre-trained model.
    \item Round 3: Using TensorFlow and the MNIST dataset, develop a convolutional neural network (CNN) to accurately classify handwritten digits. However, in this new version of the task, you must limit your training to a maximum of 500 images. Furthermore, incorporate transfer learning by using a pre-trained model to extract features from the data and improve your model's accuracy.
\end{itemize}

Example 3:
\begin{itemize}
    \item Round 0: Create a class in Java that can store a list of students and the courses they are enrolled in.
    \item Round 1: Create a Java program that manages a student database with courses, including options to add, remove, and search for students, as well as courses for each student. The program should also be able to display statistics on the number of students and courses.
    \item Round 2: Develop a Java program that manages a student database containing courses and grades. Your program should allow users to add, remove, and search for students, as well as add or drop courses for each student. Furthermore, it should display statistics on the number of students and courses, and calculate the average grade for each student. Ensure that your program implements appropriate error-handling mechanisms.
    \item Round 3: Develop a Java program to manage a library system where users can check-in, check-out and search for books in the library. The system should allow users to add new books and delete existing ones from the library. The program should also provide statistics on the number of books checked-in and checked-out, and calculate the average reading time for each book. Ensure that your program implements appropriate error-handling mechanisms and enforces data integrity constraints such as not allowing users to check-out books that have already been checked-out.
\end{itemize}

\section{Using Different Evolution Execution Models}\label{app:diff_evol}

We explored using the open-source models (OSS) CodeLlama-Instruct-34B for generating evolved instructions. However, it demonstrated relatively low coding performance in response generation. To address this, we fine-tuned it using our code-alpaca dataset and utilized this model for response generation.




\end{document}